%% file: main.tex
\documentclass[journal]{IEEEtran}
\ifCLASSINFOpdf
\else
\fi

\usepackage{cite}
\usepackage{amsmath}
\usepackage{subfig}
\usepackage{xcolor}
\usepackage{amsfonts}
\usepackage{multirow}
\usepackage{graphicx}
\usepackage{verbatim}
\usepackage{booktabs}
\usepackage{array}
\usepackage{arydshln}
\usepackage{bbding}
\begin{document}
%
% paper title
\title{A CTC Alignment-based Non-autoregressive Transformer for End-to-end Automatic Speech Recognition}
%
%
%
% author names and IEEE memberships
% note positions of commas and nonbreaking spaces ( ~ ) LaTeX will not break
% a structure at a ~ so this keeps an author's name from being broken across
% two lines.
% use \thanks{} to gain access to the first footnote area
% a separate \thanks must be used for each paragraph as LaTeX2e's \thanks
% was not built to handle multiple paragraphs
%

\author{Ruchao Fan,~\IEEEmembership{Student Member,~IEEE,}
        Wei Chu,
        Peng Chang,
        and~Abeer~Alwan,~\IEEEmembership{Fellow,~IEEE}% <-this % stops a space
\thanks{R. Fan and A. Alwan are with the Department
of Electrical and Computer Engineering, University of California, Los Angeles,
CA, 90095 USA (e-mail: fanruchao@g.ucla.edu, alwan@ee.ucla.edu).}% <-this % stops a space
\thanks{W. Chu and P. Chang are with PAII Inc., CA, USA, (email: \{chuwei129,changpeng805\}@pingan.com.cn).}
\thanks{This work is supported in part by the NSF and UCLA-Amazon Science Hub.}}
% <-this % stops a space
%\thanks{}}

% note the % following the last \IEEEmembership and also \thanks - 
% these prevent an unwanted space from occurring between the last author name
% and the end of the author line. i.e., if you had this:
% 
% \author{....lastname \thanks{...} \thanks{...} }
%                     ^------------^------------^----Do not want these spaces!
%
% a space would be appended to the last name and could cause every name on that
% line to be shifted left slightly. This is one of those "LaTeX things". For
% instance, "\textbf{A} \textbf{B}" will typeset as "A B" not "AB". To get
% "AB" then you have to do: "\textbf{A}\textbf{B}"
% \thanks is no different in this regard, so shield the last } of each \thanks
% that ends a line with a % and do not let a space in before the next \thanks.
% Spaces after \IEEEmembership other than the last one are OK (and needed) as
% you are supposed to have spaces between the names. For what it is worth,
% this is a minor point as most people would not even notice if the said evil
% space somehow managed to creep in.

% The paper headers
\markboth{IEEE Transactions on Audio, Speech, and Language Processing, 2023}{}
%\markboth{Journal of \LaTeX\ Class Files,~Vol.~14, No.~8, August~2015}%
%{Shell \MakeLowercase{\textit{et al.}}: Bare Demo of IEEEtran.cls for IEEE Journals}
% The only time the second header will appear is for the odd numbered pages
% after the title page when using the twoside option.
% 
% *** Note that you probably will NOT want to include the author's ***
% *** name in the headers of peer review papers.                   ***
% You can use \ifCLASSOPTIONpeerreview for conditional compilation here if
% you desire.

% If you want to put a publisher's ID mark on the page you can do it like
% this:
%\IEEEpubid{0000--0000/00\$00.00~\copyright~2015 IEEE}
% Remember, if you use this you must call \IEEEpubidadjcol in the second
% column for its text to clear the IEEEpubid mark.

% make the title area
\maketitle

% As a general rule, do not put math, special symbols or citations
% in the abstract or keywords.
\begin{abstract}
Recently, end-to-end models have been widely used in automatic speech recognition (ASR) systems. Two of the most representative approaches are connectionist temporal classification (CTC) and attention-based encoder-decoder (AED) models. Autoregressive transformers, variants of AED, adopt an autoregressive mechanism for token generation and thus are relatively slow during inference. In this paper, we present a comprehensive study of a CTC Alignment-based Single-Step Non-Autoregressive Transformer (CASS-NAT) for end-to-end ASR. In CASS-NAT, word embeddings in the autoregressive transformer (AT) are substituted with token-level acoustic embeddings (TAE) that are extracted from encoder outputs with the acoustical boundary information offered by the CTC alignment. TAE can be obtained in parallel, resulting in a parallel generation of output tokens. During training, Viterbi-alignment is used for TAE generation, and multiple training strategies are further explored to improve the word error rate (WER) performance. During inference, an error-based alignment sampling method is investigated in depth to reduce the alignment mismatch in the training and testing processes. Experimental results show that the CASS-NAT has a WER that is close to AT on various ASR tasks, while providing a $\sim$24x inference speedup. With and without self-supervised learning, we achieve new state-of-the-art results for non-autoregressive models on several datasets. We also analyze the behavior of the CASS-NAT decoder to explain why it can perform similarly to AT. We find that TAEs have similar functionality to word embeddings for grammatical structures, which might indicate the possibility of learning some semantic information from TAEs without a language model.
%ncluding convolutional augmented self-attention blocks for both the encoder and decoder, iterative loss, and token acoustical boundary expansion.
\end{abstract}

% Note that keywords are not normally used for peerreview papers.

% For peer review papers, you can put extra information on the cover
% page as needed:
% \ifCLASSOPTIONpeerreview
% \begin{center} \bfseries EDICS Category: 3-BBND \end{center}
% \fi
%
% For peerreview papers, this IEEEtran command inserts a page break and
% creates the second title. It will be ignored for other modes.
\IEEEpeerreviewmaketitle

\input{Tex/intro}
\input{Tex/background}
\input{Tex/methods}

\input{Tex/expsetup}

\input{Tex/resultA-C}

\input{Tex/resultD-F}

\input{Tex/conclusion}

%\appendices
%\section{Proof of the First Zonklar Equation}
%Appendix one text goes here.

%% you can choose not to have a title for an appendix
%% if you want by leaving the argument blank
%\section{}
%Appendix two text goes here.

% use section* for acknowledgment
%\section*{Acknowledgment}
%This work was supported in part by the NSF. The authors %would like to thank...

% Can use something like this to put references on a page
% by themselves when using endfloat and the captionsoff option.
\ifCLASSOPTIONcaptionsoff
  \newpage
\fi

% trigger a \newpage just before the given reference
% number - used to balance the columns on the last page
% adjust value as needed - may need to be readjusted if
% the document is modified later
%\IEEEtriggeratref{8}
% The "triggered" command can be changed if desired:
%\IEEEtriggercmd{\enlargethispage{-5in}}

% references section

% can use a bibliography generated by BibTeX as a .bbl file
% BibTeX documentation can be easily obtained at:
% http://mirror.ctan.org/biblio/bibtex/contrib/doc/
% The IEEEtran BibTeX style support page is at:
% http://www.michaelshell.org/tex/ieeetran/bibtex/
%\bibliographystyle{IEEEtran}
% argument is your BibTeX string definitions and bibliography database(s)
%\bibliography{IEEEabrv,../bib/paper}
%
% <OR> manually copy in the resultant .bbl file
% set second argument of \begin to the number of references
% (used to reserve space for the reference number labels box)
\bibliographystyle{IEEEtran}
%\bibliography{strings,refs}
\bibliography{Bib/e2e,Bib/nar,Bib/ssl}

\end{document}

%% file: Tex/intro.tex
\section{Introduction}
% The very first letter is a 2 line initial drop letter followed
% by the rest of the first word in caps.
% 
% form to use if the first word consists of a single letter:
% \IEEEPARstart{A}{demo} file is ....
% 
% form to use if you need the single drop letter followed by
% normal text (unknown if ever used by the IEEE):
% \IEEEPARstart{A}{}demo file is ....
% 
% Some journals put the first two words in caps:
% \IEEEPARstart{T}{his demo} file is ....
% 
% Here we have the typical use of a "T" for an initial drop letter
% and "HIS" in caps to complete the first word.
\IEEEPARstart{E}{nd}-to-end models have proven successful for speech recognition because of their ability to play the role of the acoustic, pronunciation, and language model in one single neural network \cite{li2020comparison, li2021recent}. Training the above components together leads to fewer intermediate errors and thus a lower word error rate (WER) for ASR systems. This training mechanism also requires fewer model parameters, which is suitable for on-device deployment. Connectionist temporal classification (CTC)\cite{graves2006connectionist}, attention-based encoder decoder (AED) \cite{chan2016listen}, and RNN-Transducers\cite{graves2012sequence, zhang2020transformer} are the most widely used end-to-end models. CTC has a high decoding efficiency when using the best path decoding strategy, but it is restricted by its assumption of conditionally independent outputs. AED, like the autoregressive transformer (AT) \cite{vaswani2017attention, dong2018speech}, models output dependencies by incorporating a language-model-style decoder. However, the decoding in AT adopts an autoregressive mechanism for joint probability factorization, leading to a step-by-step generation of output tokens. Such a mechanism lowers the inference speed for ASR, which is an essential factor when designing an efficient ASR system.

Recently, non-autoregressive mechanisms have received increasing attention for their decoding efficiency, enabled by generating output tokens in parallel \cite{gu2018non, lee2020deterministic, saharia2020non, qi2021bang, chen2020non, higuchi2020mask}. There are two major types of Non-Autoregressive methods for Transformers (NAT): (i) iterative NATs, and (ii) single-step NATs or one-shot NATs. The prevailing iterative NATs relax the strict non-autoregressive condition and iteratively generate outputs with $K$ decoding passes. Thus, iterative NATs are sometimes called ``semi-NAT''. Single-step NATs, however, can generate the output sequence in one iteration. Different from the methods in neural machine translation that extend encoder input as the decoder input, single-step NATs for speech recognition extract high-level acoustic representations as the decoder input, assuming that language semantics can be captured by the acoustic representations \cite{bai2021,tian2020spike,yu2021boundary}. These acoustic representations, however, are either implicit, extracted by attention mechanism \cite{bai2021} or incomplete using only CTC spikes \cite{tian2020spike}, which make learning language semantics difficult.
% Specifically, Bai et al. use a fixed length of decoder input to extract data-driven acoustic representation for each token \cite{bai2021}. Tian et al. use the encoder outputs that generate CTC spikes as the acoustic representation for each token \cite{tian2020spike}.

Non-autoregressive methods continue to be proposed based on CTC because of its efficiency \cite{chen21q_interspeech}. For example, Chi et al. proposed to train a refiner to iteratively improve CTC alignment based on the previous outputs of the refiner \cite{chi2020align}. In \cite{chan2020imputer}, CTC alignment is enhanced with a mask token as a prior information for the decoder in each iteration. Nozaki et al. alleviate the output-independent problem of CTC by using intermediate predictions as additional inputs \cite{nozaki21_interspeech}. Furthermore, \cite{ng2021pushing} improves the WER performance of a pure CTC model by a large margin using Wav2vec2.0 pretraining techniques. The performance of these methods, however, is still worse than their autoregressive transformer (AT) counterparts.

In this paper, we present a comprehensive study of a NAT framework by utilizing alignments over the CTC output space. The framework can generate the output sequence within one iteration, so we refer to it as CTC Alignment-based Single Step NAT (CASS-NAT). In CASS-NAT, there are four major modules: encoder, token-level acoustic embedding extractor (TAEE), self-attention decoder (SAD), and mixed-attention decoder (MAD). The encoder is used to extract a high-level acoustic representation for each frame. The TAEE extracts a more meaningful token-level acoustic embedding (TAE) using the information given by alignments over the CTC output space. The SAD and MAD model the dependencies between TAEs, where MAD considers encoder outputs directly for the purpose of source-attention while SAD does not. However, SAD indirectly uses the information from the encoder through the TAEs for self-attention. Since TAEs can be obtained in parallel, no recurrence in output sequence generation exists. Meanwhile, the two decoder modules can model the dependencies between TAEs in the latent space.

%Based on our previous CASS-NAT papers \cite{fan2021cass, fan2021improved}, in this work, we conduct sufficient ablation studies to explore the characteristics of the CASS-NAT, and thus a deeper understanding of the reason why it performs close to its autoregressive counterpart.
\textbf{We summarize the contributions of this work as: 1)} detailed experiments to examine the effect of various decoder structures on the WER, while the settings of SAD and MAD in \cite{fan2021cass} were intuitively selected; \textbf{2)} an investigation of the impact of the hyper-parameters on the proposed error-based sampling alignment (ESA) method, such as the sampling threshold, the number of sampled alignments, and the scoring model for ranking the sampled alignments. The investigation reveals a trade-off between accuracy and inference efficiency, which was not covered in \cite{fan2021cass}; \textbf{3)} comparisons of the effectiveness of each individual training strategy in \cite{fan2021improved} and their combinations are presented for a better understanding of the proposed training strategies. Knowledge distillation, which is not covered in \cite{fan2021improved}, is also included in this work; \textbf{4)} an investigation of various encoder initialization schemes (including AT encoder, CTC encoder, and random initialization) for CASS-NAT training since the quality of the CTC alignment is 
highly relevant to model accuracy. The HuBERT encoder\cite{hsu2021hubert} is included in the comparison as well. Such an investigation was not done in earlier publications; and \textbf{5)} use the proposed methods on diverse datasets (LibriSpeech: adult-read, TED2: adult-spontaneous, MyST: child-spontaneous, and Aishell1: adult-Mandarin-read) to validate the generalizability of our algorithm, and obtain new state-of-the-art ASR results for non-autoregressive models. Our earlier publications reported results on only LibriSpeech and Aishell1.

The remainder of the paper is organized as follows. Section \ref{sec:background} introduces the background of end-to-end models and related work. Section \ref{sec:cassnat} describes the CASS-NAT framework, including system architecture and training and inference strategies. Experimental setups are described in Section \ref{sec:exp_setup}, and results are shown and discussed in Section \ref{sec:results}. We conclude the paper in Section \ref{sec:conclusion}.

%% file: Tex/background.tex
\section{Background}
\label{sec:background}
We first review the important concepts behind the proposed methods and provide basic notations. Let $X=(x_1,...,x_t,...,x_T)$ denote the input sequence, where $x_t$ contains speech features of frame $t$. $Y=(y_1,...,y_u,...,y_U)$ is the output sequence, where $y_u$ is a token at position $u$. The goal of speech recognition tasks is to find the best probable transcription $Y$ given acoustic information $X$, which can be formulated as $Y^*=\underset{Y}{\mathrm{argmax}} \: P(Y|X)$.

\subsection{Autoregressive Models}
Recently, transformers are shown to be the best-performing autoregressive models (AT)\cite{dong2018speech, nakatani2019improving}. AT adopts an encoder-decoder structure, where the decoder generates each token conditioned on all previous tokens. This architecture design achieves sequence modelling by a chain of conditional probabilities, where each conditional probability constructs a classification problem. The AT model is then trained through an objective function as follows:
\begin{equation}
\label{eq:ar}
\begin{aligned}
    L_{AT} = -\log\, P(Y|X) &= -\log\, \prod_{i=1}^U P(y_i|y_{<i}, X) \\
        &= -\sum_{i=1}^U \log\, P(y_i|y_{<i}, X)
\end{aligned}
\end{equation}
where $P$ is the probability distribution of the AT model and $y_{<i}$ are all previous tokens before the $i^{th}$ token.

AT can be trained efficiently by using all ground-truth tokens as the decoder input (teacher-forcing) or using parallel scheduled sampling \cite{zhou2019improving}. During inference, however, a beam search algorithm is used to obtain the most probable sequences over the search space. The beam search and a requirement of using history tokens for generation, destroy the parallelism in AT, leading to low inference speed.  

\subsection{Non-autoregressive Models}
Non-autoregressive models have no strict dependencies between tokens. For example, CTC has a conditional independence assumption, while Mask-CTC \cite{higuchi2020mask} trains the decoder as a masked language model to build a weak dependency between masked and unmasked tokens. By relaxing the dependency assumption, it is possible to generate all tokens in parallel, and thus increase inference speed.

\subsubsection{CTC and its Alignment}
\label{sssec:ctc}
The CTC model has no dependency assumption between tokens. The posterior probability of $Y$ given $X$ can be directly computed by a multiplication of the probability for each frame with each other. Let $Z=\{z_1,...,z_t,...,z_T\}$ be the output of the model, where $z_t$ stands for the output at time step $t$ corresponding to the input $x_t$. The length of $Z$, however, is always longer than that of $Y$ for a speech recognition task. To compute the loss, a special blank token $b$ is added in the vocabulary; $b$ can be predicted as an output of $Z$ at any time step. During inference, $b$ and repeated tokens are removed to obtain a shorter sequence $Y$, where this process can be defined as a mapping rule $\beta$. During training, there exist multiple $Z$s that can be mapped to $Y$ using the rule $\beta$. As a consequence, CTC loss is a summation of all such $Z$s, which is formulated as follows \cite{graves2006connectionist}: 
\begin{equation}
\label{eq:ctc}
\begin{aligned}
    L_{CTC} &= -\log\, P(Y|X) = -\log \sum_{Z\in \beta^{-1}(Y)}P(Z|X) \\
        &=-\log \sum_{Z\in \beta^{-1}(Y)} \prod_{t=1}^T P(z_t| X)
\end{aligned}
\end{equation}
where $\beta^{-1}$ is the inverse of the mapping rule.

An aligned relationship between all time steps in $X$ and tokens in $Y$ is presented in each $Z$. Thus, $Z$ is also called an CTC alignment, which is similar to the alignment in HMM-based ASR systems. In this paper, we consider the benefits of the information offered by the alignment $Z$ to achieve a single-step NAT for end-to-end speech recognition. $Z^*$ is referred to as the Viterbi-alignment when it has the maximum probability of being mapped to the ground truth $Y$.

\subsubsection{Single-step NAT}
\label{sssec:ss_nat}
Single-step NATs use the encoder-decoder structure, making the output the same length as $Y$. The output tokens are still conditionally independent of each other, just like CTC. But there is an implicit assumption that language semantics can be captured by high-level acoustic representations, which are similar to word embeddings. The model is updated using a cross entropy loss for each token, which can be formulated as:
\begin{equation}
\label{eq:ssnat}
\begin{aligned}
    L_{NAT} = -\log\, P(Y|X) &= -\sum_{i=1}^U \log \, P(y_i | X)
\end{aligned}
\end{equation}

\subsection{Other Related Works}

\subsubsection{Convolution-augmented self-attention}
The self-attention module in transformers captures global information by a weighted summation of the whole sequence. However, local information is also important for sequence modelling. Taking speech features as an example, frequency details in each vowel or consonant helps the recognition of these sounds. In computer vision, convolution layers have proved to be good at capturing local details within a kernel \cite{bello2019attention}. Recent works adopt this idea and augment transformers with a convolution module \cite{yu2018qanet,gulati2020conformer,han2020contextnet} in ASR. Relative positional encoding is also used in each self-attention module. The convolution-augmented self-attention block can effectively improve performance, but the improvement is only significant if applied in the AT encoder. The AT decoder, on the other hand, adopts a causal structure (upper triangular mask matrix for attention) that captures less local details with a convolution module, and thus the improvement would not be significant in that case. In this paper, we explore the use of convolution-augmented self-attention layers for the NAT decoder in addition to the encoder.

\subsubsection{Intermediate Loss}
Deep transformers always suffer from gradient vanishing, especially for parameters that are distant from the output layers. Intermediate loss has previously been proposed to add additional loss functions after each layer to boost the gradient update\cite{szegedy2015going,tjandra2020deja}. It has been proven useful to use intermediate CTC loss for improving the performance of the CTC model \cite{lee2021intermediate,lee21e_interspeech}. In \cite{wang2020transformer}, intermediate CE loss is used for training deep transformer-based acoustic models for an HMM-based hybrid ASR system. In this paper, we combine the usage of intermediate CTC and CE loss in the proposed framework.

\subsubsection{Self-supervised Learning}
Self-supervised learning (SSL) has been popular in recent years, especially for low-resource tasks\cite{baevski2020wav2vec, hsu2021hubert}. We use HuBERT in our experiments to further improve the system performance. During pretraining, the HuBERT model reconstructs the masked input given unmasked areas. We take advantage of self-supervised pretraining to improve the modelling capability of the encoder, and thus the quality of the CTC alignment, which is important for accurate language semantics modelling in the NAT decoder.

%There are two different types of SSL methods: future predictive loss and mask reconstruction loss. For predictive loss, the model tries to predict the future frames with its history, such as APC \cite{chung2019unsupervised} and CPC, where CPC considers negative samples \cite{oord2018representation}. As for reconstruction loss, the model reconstructs the masked input given unmasked areas, such as Tera, MPC, Wav2vec2.0, and Hubert \cite{liu2020mockingjay, liu2021tera, wang2020transformer, jiang2021further, baevski2019effectiveness, baevski2020wav2vec, hsu2021hubert}. Techniques like vector quantization, contrastive learning, k-means clustering are considered in these methods to achieve better pretraining. We take advantage of self-supervised pretraining to improve modelling capability of the encoder, and thus the quality of the CTC alignment, which is important for accurate language semantics modelling in the NAT decoder.

%% file: Tex/methods.tex
\section{Proposed Framework: CASS-NAT}
\label{sec:cassnat}
In this section, we introduce the proposed CTC Alignment-based Single Step Non-autoregressive Transformer (CASS-NAT). We describe the mathematical derivation of the objective function, training strategies to improve performance, and the proposed sampling-based decoding strategy.

\begin{figure}[tp]
\centering
\centerline{\includegraphics[width=0.48\textwidth]{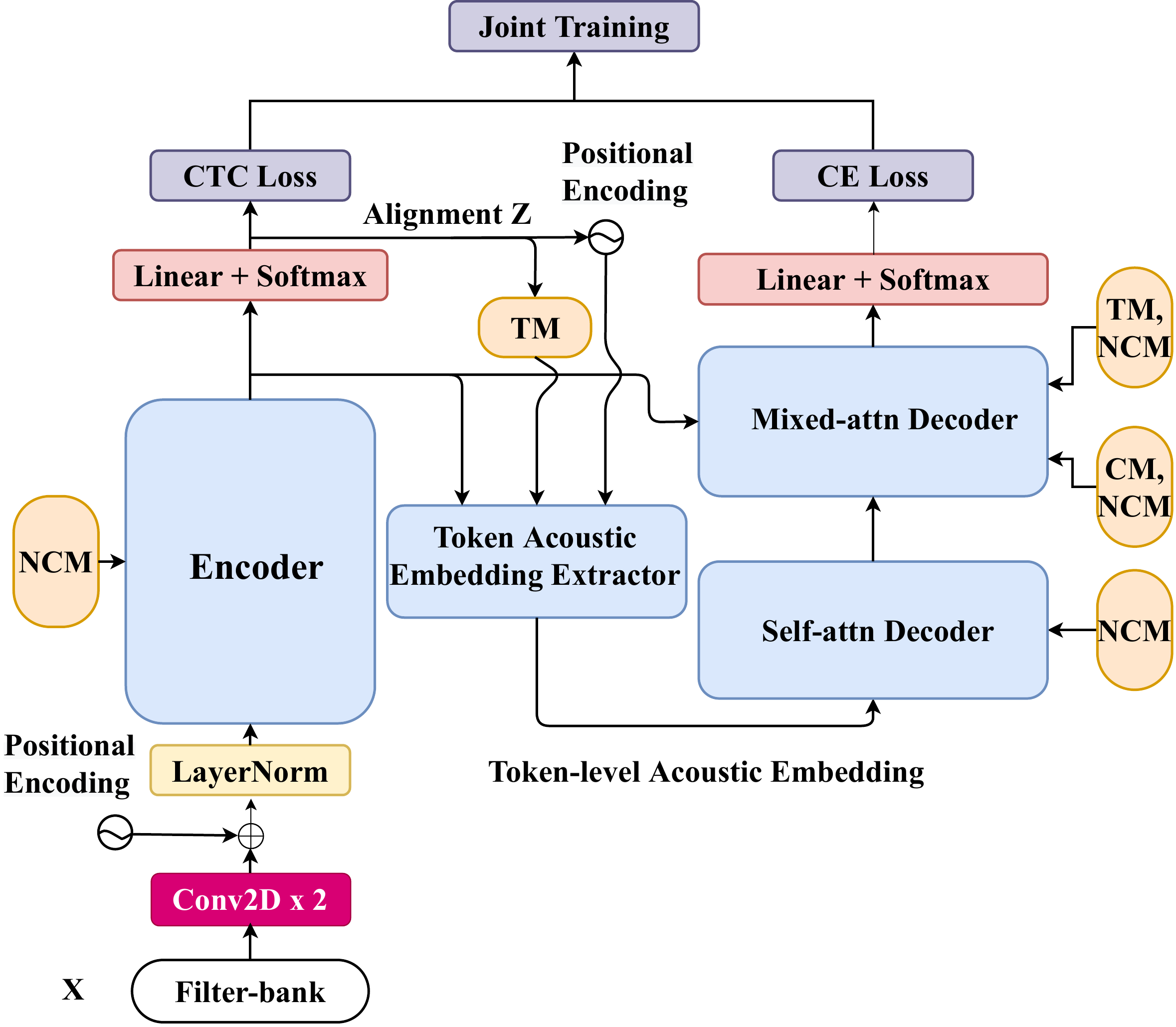}}
\caption{An overview of the proposed CASS-NAT architecture. CM and NCM represent a causal and non-causal mask, respectively. TM stands for trigger mask.}
\label{fig:nast}
\end{figure}

\subsection{System Architecture}
\label{ssec:model}
% model -> system architecture
The proposed CASS-NAT system architecture builds upon the CTC/Attention hybrid architecture\cite{watanabe2017hybrid} to be non-autoregressive using CTC alignments. Fig.\ref{fig:nast} shows the four major modules in CASS-NAT: encoder, token acoustic embedding extractor (TAEE), self-attention decoder (SAD) and mixed-attention decoder (MAD).

%\hspace*{\fill} 
\subsubsection{\textbf{Mask in attention}}
\label{sssec:mask}
The attention mechanism is important for a transformer. The most basic computation in the attention mechanism is scaled dot-product self-attention using a sequence as input. Conventional self-attention utilizes information across the whole sequence. But each output of the self-attention mechanism is not necessarily dependent on all of the input sequence. For example, a neural language model uses an upper triangular matrix (mask matrix) to gather information from only past tokens in a sequence. In Fig.~\ref{fig:nast}, we show three mask matrices for different purposes in the four major modules. Since we do not consider a streaming ASR model at this moment, non-causal mask (NCM) is used in the encoder. The NCM is a matrix where the paddings are zeros to prevent the padded tokens or padded frames from attention computation. In TAEE, a trigger mask (TM) is used to extract accurate token-level acoustic embeddings. TM marks out the triggered frames, such that the positions of used frames are marked as ones, while other positions are marked as zeros. Examples of TM can be found in Section \ref{sssec:TAE}. MAD contains both self-attention and cross-attention layers, which are similar to an AT decoder. In such cases, either a causal mask (CM) or NCM can be used for the self-attention layer, and either a NCM or TM can be used for the cross-attention layer. The different choices of the mask matrices in MAD are explored experimentally in Section \ref{ssec:decoder_structure}. Eventually, the self-attention computation can be augmented with a mask matrix as follows:
\begin{equation}
\label{eq:attention}
    Attention(Q, K, V, M) = \left(Softmax\left( \frac{QK^T}{\sqrt{d_k}}\right) \otimes M\right) V
\end{equation}
where $Q \in R^{n_q\times d_q}$, $K \in R^{n_k\times d_k}$, $V \in R^{n_v\times d_v}$, and $M \in R^{n_q \times n_k}$ are the query, key, value and mask matrices, respectively.

%\hspace*{\fill} 
\subsubsection{\textbf{Encoder}}
\label{sssec:encoder}

The encoder extracts high-level acoustic representations $H$ from speech features $X$. A linear layer with a CTC loss function is added after the encoder as shown in Fig.\ref{fig:nast}. The role of CTC is to obtain an alignment over the CTC output space to offer auxiliary information for the token acoustic embedding extractor (TAEE). During training, Viterbi-alignment is used. During inference, various methods for sampling from the CTC output space are explored experimentally.

%Specifically,  The segment can be converted to the trigger mask in Fig. \ref{fig:nast}, which is used for self-attention computation. The information obtained from the alignment is then used to extract token-level acoustic embeddings that replace word embeddings which exists in AT. 

%The self-attention and mixed-attention decoders are finally used to model the relationship between tokens, where the MAD has access to the acoustic representations $H$, while the SAD does not.

%\hspace*{\fill} 
\subsubsection{\textbf{Token Acoustic Embedding Extractor}}
\label{sssec:TAE}
The token acoustic embedding extractor is designed to extract token-level acoustic embedding (TAE) with the auxiliary information offerred by the CTC alignment. For example, given an alignment $Z=\{z_1,...,z_t,...,z_T\}$, we can estimate an acoustic segment for each token $u$ as $\{t_{u-1}+1,...,t_u\}$ (note that 1 here refers to one frame), and the number of tokens in $Z$.

First, CTC alignments offer an acoustic boundary for each token $y_u$, which is then transformed into the trigger mask. Specifically, we define a mapping rule from alignment to trigger mask, and fix the rule in both the training and inference phases. We regard the first non-blank index of each token in the alignment as its end boundary. The intuition is our assumption that the model will not output a token until it sees all the acoustic information.of the token. Using the first non-blank index is for simplicity and consistency in training and decoding. For example, if an alignment is $Z=\{\_, C, C, \_, A, \_, \_, T, \_\}$ for the ground truth $Y=\{C, A, T\}$, where $\_$ is the blank symbol, the end boundary for $C$ and $A$ is $Z_2$ and $Z_5$, respectively, and thus the trigger mask for token $A$ is $[0,0,1,1,1,0,0,0,0]$. The mapping rule might not be accurate for acoustic segmentations, but it should be consistent during the training and decoding.
% from $z_1$ to $z_2$ and so forth for other tokens without overlap%
The trigger mask here is different from that used in \cite{moritz2020streaming} for streaming purposes. In \cite{moritz2020streaming}, previous acoustic representations could be reused for each token, and thus the trigger mask in the previous example for token $A$ is $[1,1,1,1,1,0,0,0,0]$. 

Second, CTC alignments provide the number of tokens for the decoder input. After removing blank symbols and repetitions, the number of tokens in an alignment $Z$ is used as the predicted length of sinusoidal positional encoding (decoder input length). As shown in Fig.~\ref{fig:nast}, TAE for each token is then extracted with the trigger mask and the sinusoidal positional encoding using a one-layer source-attention block. The TAEs replace the word embeddings in AT to achieve parallel generation of each sequence.

%\hspace*{\fill} 
\subsubsection{\textbf{Self-attention Decoder}}
\label{sssec:sad}
TAEs extracted from TAEE (see Section \ref{sssec:TAE}) have the good property of parallel generation and thus is used as a substitution of word embeddings for the decoder input. Since there is no need to create recurrence in the decoder, we use a non-causal mask (NCM) in the self-attention decoder (SAD) to model the relationships between TAEs.

%\hspace*{\fill} 
\subsubsection{\textbf{Mixed-attention Decoder}}
\label{sssec:mad}
We assume that TAE has a similar capability of learning language semantics compared to the word embedding. Hence, we design a mixed-attention decoder (MAD) to retrace the encoder information for better decision making at the output layer. Similar to an AT decoder, MAD has a self-attention layer that uses either CM or NCM, and a source-attention layer that uses either TM or NCM. A linear layer is added after MAD, followed by a cross-entropy loss. Since we use the Viterbi-alignement during training, the output has the same length as the ground truth $Y$. 

\subsection{Training Details}
\label{ssec:training}
The training criterion is presented in this section, followed by various training strategies used to improve the performance of CASS-NAT.
%Different from the work where latent variable NAT was proposed for neural machine translation \cite{shu2019latent}, we don't use any additional networks for distribution estimation and length prediction.

%\hspace*{\fill} 
\subsubsection{\textbf{Training Criterion}}
In our framework, CTC alignments $Z$ are introduced as latent variables. Given $X$ and $Y$ in Sec.\ref{sec:background}, the log-posterior probability can be decomposed into:
\begin{equation}
\label{eq:decompose}
\begin{aligned}
 %\small
 \log \, P(Y|X) = \log \, \mathbb{E}_{Z|X}[P(Y|Z,X)],  \quad Z \in q.
%L_{\text{transformer}} = \max_Z{\log P(Y|Z,X)}},  \quad Z \in q,
\end{aligned}
\end{equation}
where $q$ is the set of alignments that can be mapped to $Y$. For those alignments that do not belong to $q$, we assume that $P(Y|Z, X)$ is equal to zero. To reduce computational cost, the maximum approximation \cite{zeyer2020new} is applied:
\begin{equation}
\label{eq:maxapprox}
\begin{aligned}
   %\small
   \log \, P(Y|X) &\geq \mathbb{E}_{Z|X}[\log\, P(Y|Z,X)] \\
   &\approx \max_Z{\log\,\prod_{u=1}^U{P(y_u|z_{t_{u-1}+1:t_{u}}, x_{1:T})}}
%L_{\text{transformer}} = \max_Z{\log\prod_{u=1}^U{P(y_u|z_{t_u:t_{u+1}}, x_{1:T})}}
\end{aligned}
\end{equation}
where $\mathbb{E}$ represents the expectation and $t_{u}$ is the end boundary of token $u$ ($t_0=0$). All $t_u$s can be estimated from the alignment $Z$. We expect that TAEs can capture language semantics to a certain degree (see Section \ref{ssec:why_cassnat} for analysis), which helps alleviate performance degradation caused by the independence of the output tokens.

The framework is trained by jointly optimizing a CTC loss function on the encoder side ($L_{CTC}$) and a cross entropy (CE) loss function on the decoder side ($L_{dec}$) with a task ratio $\lambda$ \cite{watanabe2017hybrid}, and thus the final loss function ($L_\text{joint}$) is defined as:
\begin{equation}
    %\scriptsize
    L_{\text{joint}} = -\lambda \cdot \log \sum_{Z\in q}{\prod_{i=1}^T{P(z_i|X)} - \log\prod_{u=1}^U{P(y_u|z^{*}_{t_{u-1}+1:t_{u}}, X)}}
\label{eq:cassloss}
\end{equation}
where P is the probability distribution, $Z^*$ is the most probable alignment (Viterbi-alignment). The second term is a maximum approximation for the log-posterior probability as computed by Eq. \ref{eq:maxapprox}.

%\hspace*{\fill} 
\subsubsection{\textbf{Convolution Augmented Self-attention Block}}
The self-attention computation in Eq.\ref{eq:attention} considers global information across the sequence, but ignores local details. To alleviate this problem, convolution augmented self-attention blocks are proposed to emphasise the modelling of local dependencies of the input sequence in the encoder \cite{gulati2020conformer,yang-etal-2019-convolutional}. Different from previous work, we apply the convolution augmented self-attention blocks in the SAD and MAD as well. Specifically, the feed-forward layer is decomposed into two sub-layers to be placed at the beginning and the end of the block. A convolution layer similar to that in \cite{gulati2020conformer} is inserted after the self-attention layer except that we empirically use layer normalization instead of batch normalization. The final computation in the $i^{th}$ MAD can be formulated as:

\begin{align}
\label{eq:mad_block}  
    \hat{s_i} &= s_i + \frac{1}{2}\text{FFN}(s_i) \\
    s_i^{'} &= \hat{s_i} + \text{LN}(\text{Attn}(\hat{s_i}, \hat{s_i}, \hat{s_i}, \text{NCM})) \\
    s_i^{''} &=  s_i^{'} + \text{Conv}(s_i^{'}) \\
    s_i^{'''} &= s_i^{''} + \text{LN}(\text{Attn}(s_i^{''}, H, H, \text{NCM})) \\
    o_i &= \text{LN}(s_i^{'''} + \frac{1}{2}\text{FFN}(s_i^{'''}))
\end{align}
where LN indicates layer normalization and FFN is the feed-forward network. NCM is non-causal mask. $s_i$ and $o_i$ are the input and output of block $i$, respectively. The convolution-augmented self-attention block can be used for other NAT models.

Different from the usage of relative positional encoding in \cite{dai2019transformer,gulati2020conformer}, we consider a maximum length of the relative position k as in \cite{zhou2019improving}. Therefore, $2k+1$ position embedding are learned to represent the relative position between $[-k, k]$.

%\hspace*{\fill} 
\subsubsection{\textbf{Intermediate Loss}}
Since CTC and CE loss functions are jointly optimized in the CASS-NAT framework, we incorporate intermediate CTC and intermediate CE loss functions into Eq.\ref{eq:cassloss} so that the parameters in different layers can be updated at the same scale. Let $L_{dec}=-\log\prod_{u=1}^U{P(y_u|z^{*}_{t_{u-1}+1:t_{u}}, X)}$ and $L_{CTC}=-\log \sum_{Z\in q}{\prod_{i=1}^T{P(z_i|X)}}$, the objective function is re-written as:
\begin{equation}
\begin{aligned}
    %\scriptsize
    L_{\text{joint}} &= \lambda_{CE}L_{dec}^{final} + (1-\lambda_{CE})L_{dec}^{mid} \\
    &+\lambda_{CTC}L_{CTC}^{final} + (1-\lambda_{CTC})L_{CTC}^{mid}
\end{aligned}
\end{equation}
where $\lambda_{CE}$ and $\lambda_{CTC}$ are task ratios. $mid$ and $final$ indicate the layer position of the inserted loss functions. We found intermediate loss to be more effective for CASS-NAT than AT models \cite{fan2021improved}. The intermediate loss can be added to other NAT models as well.

%\hspace*{\fill} 
\subsubsection{\textbf{Trigger Mask Expansion}}
The quality of TAE relies on the accuracy of the trigger mask (TM), which is mapped from the CTC alignment. Although the CTC loss function is used to optimize the alignment, there are still errors when doing forced alignment over the CTC output space, leading to an inaccurate TM. To address this issue, we expand TM to include contextual frames for each token. For example, suppose the contextual frame size is one, the acoustic boundary of token U becomes $\{z_{t_{u-1}},...,z_{t_u+1}\}$. The trigger mask will then be expanded by one in the subsequent acoustic embedding extraction. Note that trigger mask expansion is designed specifically for CASS-NAT.

\subsection{Inference: Error-based Sampling Decoding}
\label{ssec:esa_decoding}
\begin{figure}[tp]
\centering
\centerline{\includegraphics[width=0.48\textwidth]{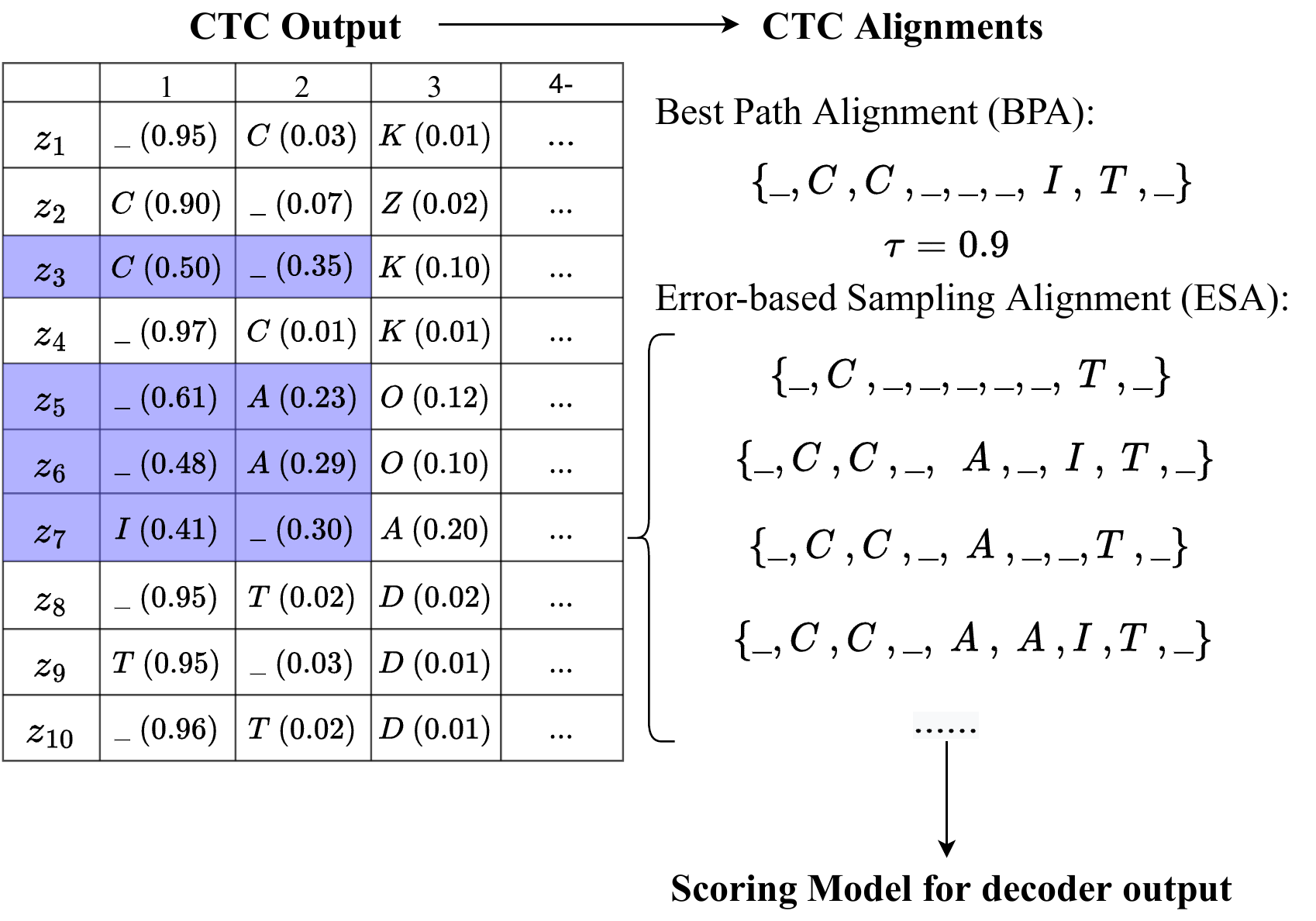}}
%  \vspace{1.5cm}
\caption{Illustration of obtaining CTC alignments from CTC outputs, including best path alignment (BPA) and error-based alignment sampling alignment (ESA). $C(0.90)$ indicates $P(z_i=C|X)=0.90$. ``$\_$" stands for a blank token. The threshold $\tau$ for sampling is set to 0.9.}
\label{fig:esa_decode}
\end{figure}

During decoding, it is essential to obtain a CTC alignment that is close to the hypothetical Viterbi-alignment used in training. The transcription, however, is not available. We propose to use three different alignment generation methods for inference: (1) best path alignment (BPA); (2) beam search alignment (BSA); and (3) error-based sampling alignment (ESA). We also present the results of using Viterbi-alignment as an upper bound of WER, assuming that transcriptions are available during decoding, which is referred to as oracle alignment. \textbf{BPA} is similar to CTC greedy decoding that selects the token with the highest probability at each time step, but without removing blank and repetitive tokens in the final sequence. \textbf{BSA} is similar to beam search decoding over the CTC output space, which is the most probable alignment during decoding.
Compared to BPA, BSA is supposed to generate an alignment that is closer to the oracle alignment, which could lead to a lower WER, but the parallelism of the CASS-NAT would be destroyed, resulting in a significant increase of the real time factor (RTF). 

Considering the expectation in Eq. \ref{eq:maxapprox}, we propose a third alignment generation approach by sampling based on the potential errors in BPA. The method is referred to as error-based sampling alignment (ESA). To generate ESA, a threshold $\tau$ determines whether sampling is required at each time step. If the highest probability of the output distribution is larger than $\tau$, the rule of BPA holds. Otherwise, we randomly sample from the tokens with the first two largest probabilities. As shown in shaded blue in Fig.~\ref{fig:esa_decode}, CTC outputs at $z_3$, $z_5$, $z_6$, and $z_7$ need a sampling because of their low top-1 probability. According to the proposed sampling rule, four sampled alignments are shown as examples on the right side of Fig.~\ref{fig:esa_decode}. The reason of sampling within the top-2 tokens is that the trigger mask is sensitive to blank tokens and most mistaken outputs in BPA contain blanks in the top-2 tokens. In addition, sampling within 2 tokens is efficient because of the small sampling space. ESA aims at correcting the output which is prone to errors. Sampling in the decoding stage will not affect much inference speed because it can be done in parallel. It is possible that ESA-generated alignments are closer to the oracle alignment than BPA. Compared to BSA, ESA can be implemented in parallel, avoiding any increase in the RTF. Finally, either an AT baseline or language model can be used to score and identify the best overall alignment.
Note that ESA can have different lengths for decoder input compared to BPA. As shown in Fig.~\ref{fig:esa_decode}, the decoder length is 4 for BPA (including an EOS token), while the lengths are 3, 5, 4, and 5 in the case of ESA. This fluctuation of the token numbers allows ESA to possibly sample an alignment that is of the same length as the oracle alignment, which is important for the performance of NAT models as will be shown experimentally. Note that ESA decoding is proposed specifically for CASS-NAT.

%% file: Tex/expsetup.tex
\section{Experimental Setup}
\label{sec:exp_setup}

\subsection{Data Preparation}
\label{ssec:data}

To examine the effectiveness of the proposed framework, we conduct several ASR tasks, including read and spontaneous speech, English and Mandarin speech, and adult and child speech. Four datasets are selected: (1) the 960-hour LibriSpeech English corpus\cite{panayotov2015librispeech} with read speech, (2) the 178-hour Aishell1 Mandarin corpus\cite{bu2017aishell} with adult read speech, (3) the 210-hour TEDLIUM2 (TED2) English corpus\cite{rousseau2014enhancing} with TED talk speech, and (4) My Science Tutor (MyST) Kids English corpus \cite{ward2011my} with spontaneous speech. We use the annotated part of MyST, which accounts for 42\% (240 hours) of the corpus.

All experiments use 80-dim log-Mel filter-bank features, computed every 10ms with a 25ms Hamming window. Features of every 3 consecutive frames are concatenated to form a 240-dim feature vector as the input. The sets of output labels consist of 5k word pieces for LibriSpeech, and 500 word pieces for TED2 and for MyST. All sub-words are obtained by SentencePiece\cite{kudo2018sentencepiece} using the training set of each dataset. 4230 Chinese characters are used as vocabulary for the Aishell1 dataset. To avoid overfitting, we applied speed perturbation\cite{ko2015audio} and SpecAugment\cite{park2019specaugment} to the filter-bank features from TED2, Aishell1 and MyST. Speed perturbation is not used for LibriSpeech because of limited computational resources.

\subsection{Network Architecture}
\label{ssec:modelarc}

\subsubsection{Autoregressive Models}
A CTC/Attention AT baseline is first trained with the architecture ($N_e=12$, $N_d=6$, $d_{ff}=2048$, $nh=8$, $d_{att}=512$) for LibriSpeech, and ($N_e=12$, $N_d=6$, $d_{ff}=2048$, $nh=4$, $d_{att}=256$) for the other three datasets, where $d_{ff}$ is the dimension of the FFN module, $d_{att}$ stands for the dimension of the attention module, $nh$ is the number of attention heads, $N_e$ and $N_d$ are the number of encoder and decoder blocks, respectively. Prior to the encoder, two convolution layers with 64 filters, a kernel size of $3\time 3$, and a stride of 2 is adopted, leading to a 4x frame-rate reduction. When using a conformer structure to the AT encoder, we reduce the $d_{ff}$ to be 1024 to keep the number of parameters in the model the same as in the transformer baseline, and the maximum length of relative position k is set to 20. The kernel size in the convolution module is 31 for LibriSpeech and 15 for the other three datasets.

\subsubsection{CASS-NAT}
\label{sssec:expcass_nat}
During training, CASS-NAT encoder is initialized with an AT encoder for faster convergence as in \cite{fan2019online}. The decoder in AT baseline is replaced by 1-block TAEE, m-block SAD and n-block MAD. $m$ and $n$ are investigated using the LibriSpeech dataset, and the best setting is applied to the other three datasets. For convolution-augmented decoder, the dimension of feed-forward layers is also halved and the maximum length of relative position is set to 8 for the tasks with word piece units and 4 for the Aishell1 data. The contextual frame of trigger mask expansion is set to 1 because we do not see further improvements with a larger expansion. The intermediate loss functions are inserted in the middle layer of the encoder and MAD with $\lambda_{CE}$ of 0.99 and $\lambda_{CTC}$ of 0.5. The inserted projection layers are discarded during inference. These settings were chosen empirically.

\subsection{Training and Decoding Setup}
\label{ssec:exp_train}
All experiments are implemented with Pytorch \cite{paszke2019pytorch} \footnote{Our code will be available at https://github.com/Diamondfan/cassnat\_asr}. The Double schedule in \cite{park2019specaugment} is adopted as the learning schedule for the LibriSpeech and Aishell datasets, where the learning rate is ramped up to and held at 0.001, then be exponentially decayed to 1e-5. The transformer-based scheduler in \cite{vaswani2017attention} with warm-up steps of 25k and noam factor of 5 is used for the TED2 and MyST datasets. Layer normalization, dropout with rate of 0.1 and label smoothing with a penalty of 0.1 are all applied as the common strategies for training a transformer. We compute WERs of development sets for early stopping (no improvement for 11 epochs). The last 12 epochs are averaged for final evaluation. Most experiments end within 90 epochs. In order to investigate the impact of different models for scoring alignments in ESA, a transformer-based language model is trained with the provided text in LibriSpeech. The provided n-gram models in the dataset are also compared in the experiments. In terms of the pretrained HuBERT encoders, we use the Fairseq model that is trained from LibriSpeech 960-hour corpus for finetuning the English data model and the Tecent model that is trained from 10000-hour WenetSpeech \cite{zhang2022wenetspeech} for finetuning the Mandarin data model. 

During AT decoding, the beam size is set to 20 for LibriSpeech, and 10 for the other three datasets. No external language models are used during beam search decoding. The evaluation of the real time factor (RTF) is conducted using a V100 GPU with a batch size of one. For CASS-NAT decoding with ESA, the threshold, number of sampled alignments and scoring models are investigated in Section \ref{ssec:esa_detail}.

%% file: Tex/resultA-C.tex
\section{Results and Analyses}
\label{sec:results}
The first set of ASR experiments used the LibriSpeech dataset to explore various decoder structures, impact factors in ESA decoding, CTC alignment behaviour, and the effectiveness of the proposed training strategies. Using the best settings found with the LibriSpeech task, experiments are run with the other three datasets.

%----------------------------------------------------
%----------------------------------------------------

\subsection{The Structure of CASS-NAT Decoder}
\label{ssec:decoder_structure}

\begin{table}[t]
\caption{WERs for different block numbers of the self-attention decoder (SAD) and mixed-attention decoder (MAD) using LibriSpeech. Various mask matrices used in MAD are also explored, where CM, NCM and TM represent causal mask, non-causal mask and trigger mask, respectively. For example, NCM + TM indicates that NCM is used for self-attention in MAD and TM is used for source-attention in MAD. Bold numbers represent the best results.}
%\scriptsize
%\footnotesize
\centering
\begin{tabular}{c c cccc}
\hline
\multirow{2}{*}{\shortstack{Decoder \\ Structure}} & \multirow{2}{*}{\shortstack{Mask in \\MAD}} & dev- &  dev- &  test- &  test- \\
~ & ~ & clean &  other &  clean &  other \\
\hline\hline
\rule{0pt}{2ex} 
7SAD + 0MAD & - & 5.3 & 11.1 & 5.4 & 11.1 \\
\hdashline
\rule{0pt}{2ex}
\multirow{3}{*}{5SAD + 2MAD} & NCM + NCM & \textbf{4.7} & \textbf{10.4} & \textbf{4.8} & \textbf{10.3} \\
~ & NCM + TM & 4.7 & 10.5 & 4.9 & 10.5  \\
~ & CM + NCM & 4.8 & 10.7 & 4.9 & 10.6  \\
\hdashline
\rule{0pt}{2ex}
3SAD + 4MAD & NCM + NCM & 4.7 & 10.4 & 4.8 & 10.4 \\
1SAD + 6MAD & NCM + NCM & 4.6 & 10.5 & 4.7 & 10.4 \\
\hline
\end{tabular}
\label{tab:decoder}
\end{table}

To investigate the structure of the CASS-NAT decoder, we use various combinations of $m$ SAD blocks and $n$ MAD blocks ($m+n=7$) in the decoder to keep the number of model parameters similar to that of the AT baseline. We use best path alignment during inference and discuss other decoding strategies in the next section. Results are shown in Table \ref{tab:decoder}. As shown in the table, not using MAD (0 MAD in the first row) causes a big performance degradation compared to other configurations, indicating that token-level speech representations might have to retrace the fine-grained frame-level information (encoder outputs) for better contextual modelling. The best performance is achieved when using 5 SADs and 2 MADs (WER of 10.3\% on the test-other set).

Because there are two attention layers (self-attention and source-attention) in each MAD, we try different mask matrices for the two layers. For the self-attention layer, either a causal mask (CM) or non-causal mask (NCM) is considered. For the source-attention layer, either a trigger mask (TM) or NCM is used. The results in Table \ref{tab:decoder} show that using NCM on both attention layers results in the best WER. Although TAE is regarded as a substitution of word embedding, CM seems not necessarily required in the CASS-NAT decoder for contextual modelling of token-level acoustic embeddings. The settings that achieve the best performance in this section (5 SADs and 2 MADs with NCM in both attention layers) are selected as default for subsequent experiments. 

%----------------------------------------------------
%----------------------------------------------------

\subsection{Error-based Sampling Alignment (ESA) Decoding}
\label{ssec:esa_detail}
\begin{figure}[t]
    \centering
    \subfloat[Effect of $P$ using test-clean]{{\includegraphics[width=0.25\textwidth]{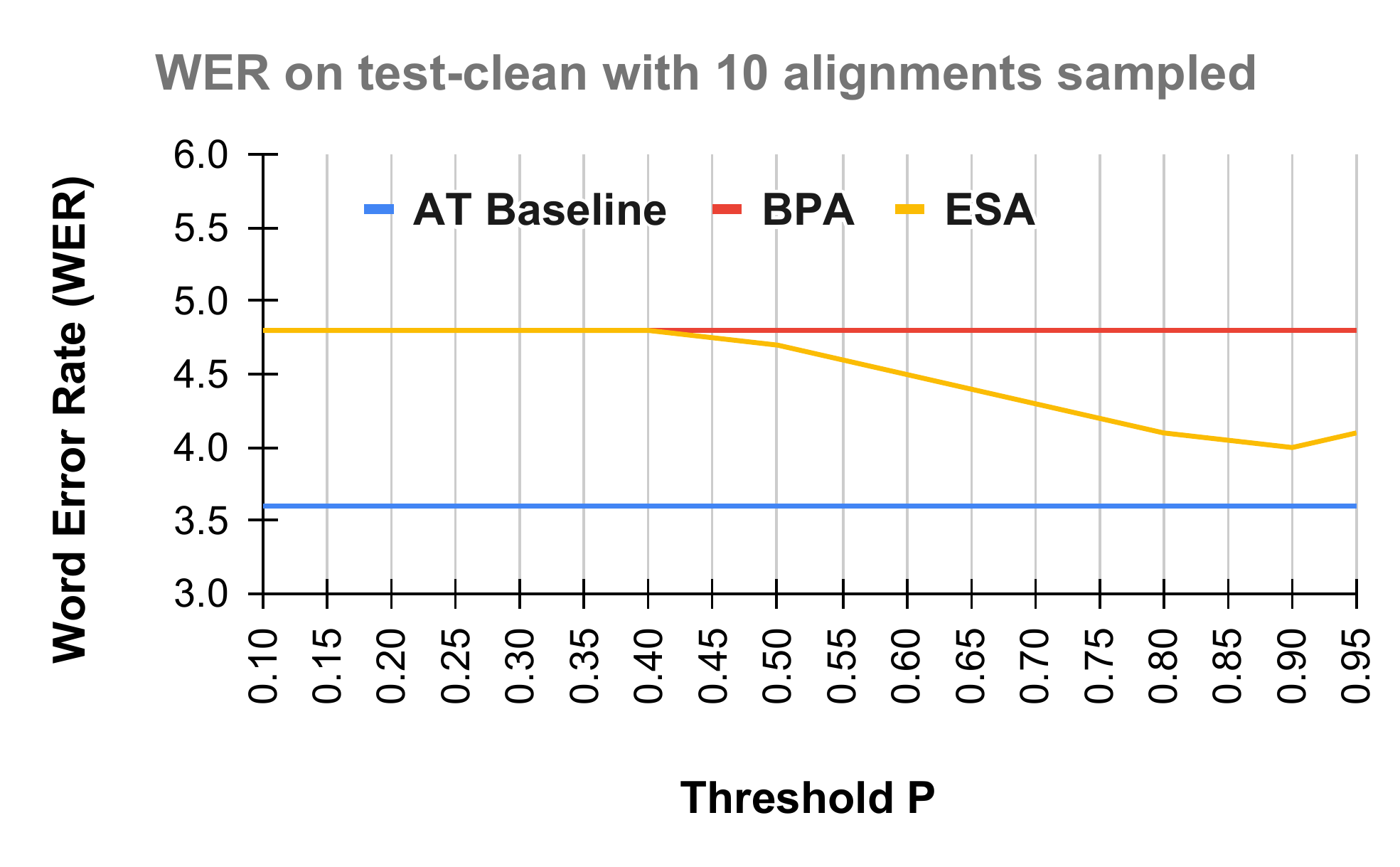} }}%
    \subfloat[Effect of $P$ using test-other]{{\includegraphics[width=0.25\textwidth]{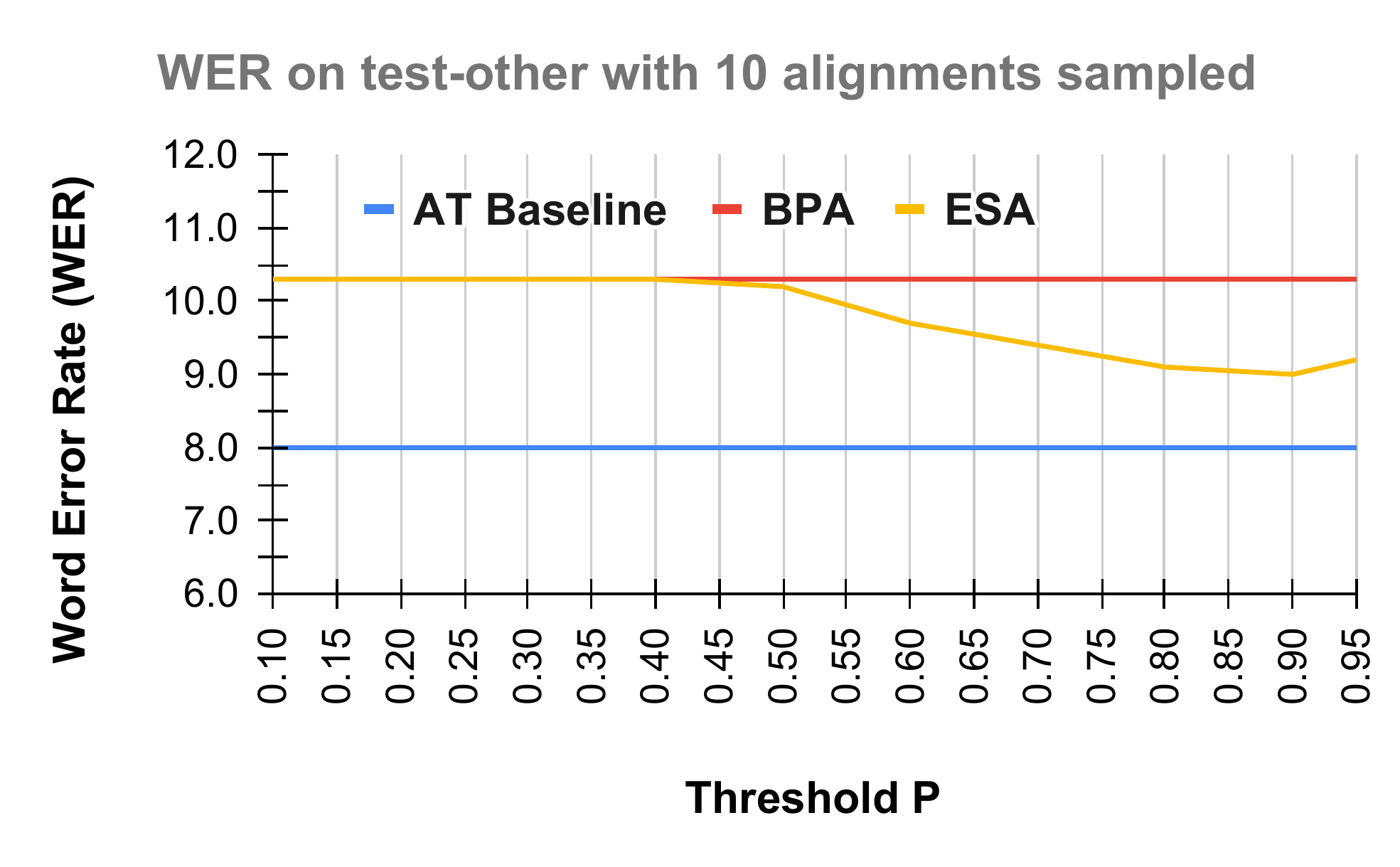}}}%
    \qquad
    \subfloat[Effect of the number of sampled alignments $S$]{{\includegraphics[width=0.42\textwidth]{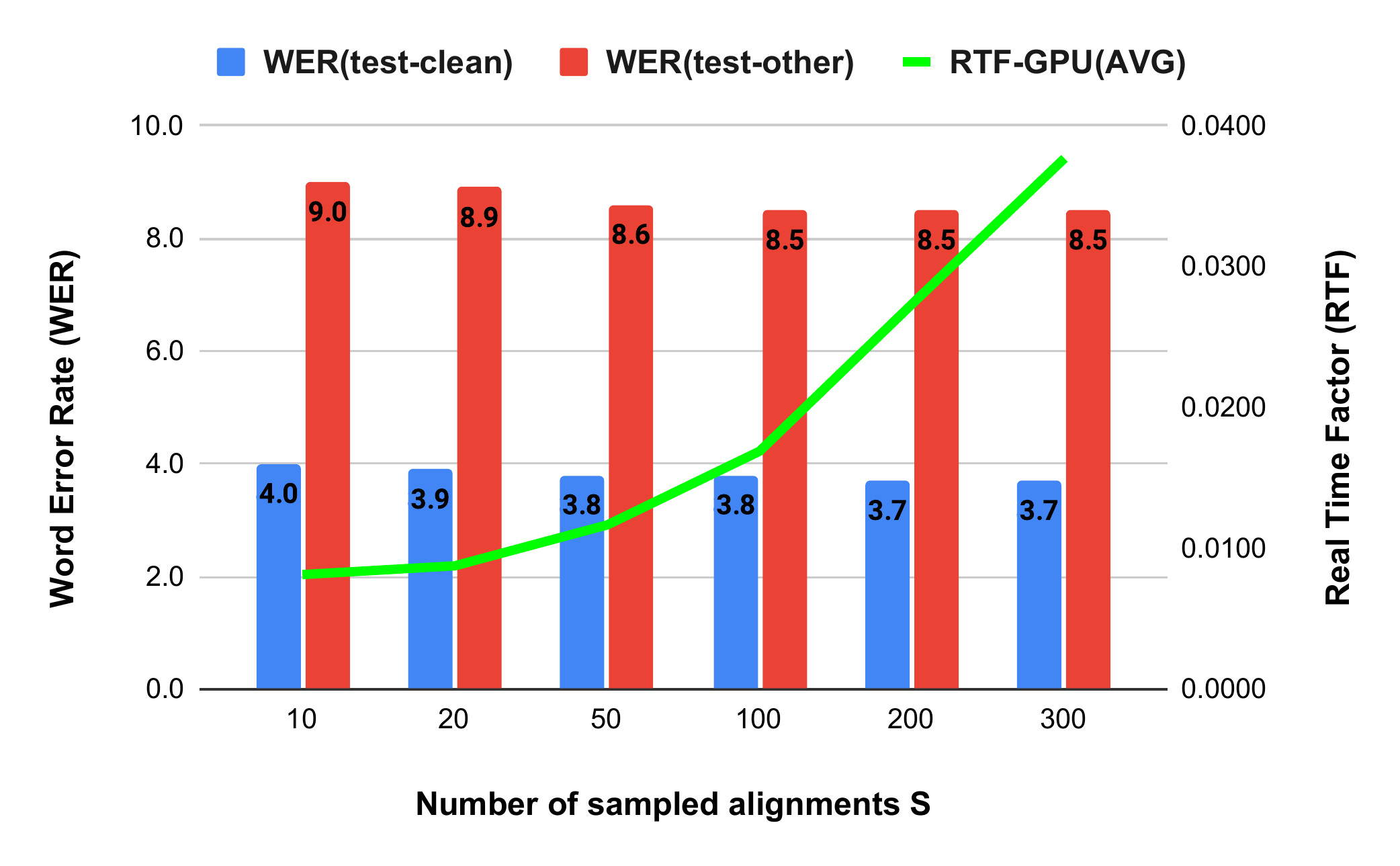}}}%
    \caption{WER performance of different values of the threshold $P$ and the number of sampled alignments $S$ in error-based sampling alignment (ESA) decoding. Real time factor (RTF) is evaluated using a V100 GPU with a batch size of one.}%
    \label{fig:esa}
\end{figure}

Viterbi-alignment is used in training, but not available during inference. To reduce the alignment mismatch between training and inference, we propose error-based sampling alignment (ESA). There are three important factors that can affect the WER performance of ESA: sampling threshold $P$, the number of sampled alignments $S$
and scoring model for ranking alignments. We use the best decoder structure in the previous section (5SAD + 2MAD) to evaluate the performance of ESA decoding with different configurations.

Figures \ref{fig:esa}a and \ref{fig:esa}b show the results when the threshold $P$ varies from 0.10 to 0.95 using the LibriSpeech test-clean and test-other data, respectively. The number of sampled alignments here is 10 and the scoring model is the AT baseline. We also include the WERs of the AT baseline and CASS-NAT decoding with best path alignment (BPA) as comparisons. As shown in the figures, ESA reaches the best performance when $P=0.9$. A higher threshold indicates fewer sampled alignments, and thus no further improvement is observed. The threshold $P$ is set to 0.9 in subsequent experiments.

Fig.\ref{fig:esa}c shows the effect of the number of sampled alignments $S$ in terms of WER and RTF on the test-clean and test-other data. As observed in the figure, by increasing the number of sampled alignments, the WER of ESA decoding improves but the improvement is small when $S$ is greater than 50. Meanwhile, the RTF increases rapidly as $S$ increases. Overall, $S=50$ might be the most appropriate value to use as default in subsequent experiments.

\begin{table}[t]
\caption{Comparisons, in terms of WER and GPU speedup, of BPA, BSA, and ESA decoding. NLM is the transformer-based neural language model. 3-gram and 4-gram models are the ones offered in LibriSpeech.}
\scriptsize
%\footnotesize
\centering
\begin{tabular}{c c cccc c}
\hline
\multirow{2}{*}{} & \multirow{2}{*}{\shortstack{Scoring \\Model}} & dev- &  dev- &  test- &  test- & \multirow{2}{*}{\shortstack{GPU \\Speedup}} \\
~ & ~ & clean &  other &  clean &  other & ~ \\
\hline\hline
\rule{0pt}{2ex} 
AT baseline & - & 3.4 & 8.1 & 3.6 & 8.0 & 1.00x \\
\hline
Oracle & - & 2.1 & 5.5 & 2.2 & 5.3 & 39.6x\\
BPA & - & 4.7 & 10.4 & 4.8 & 10.3 & 90.0x \\
BSA & - & 3.8 & 8.8 & 3.9 & 8.8 & 0.90x \\
\hdashline
\rule{0pt}{2ex}
\multirow{4}{*}{ESA} & AT baseline & 3.6 & 8.8 & 3.8 & 8.6 & 28.4x \\
~ & NLM & 3.6 & 8.9 & 3.9 & 8.7 & 31.6x \\
~ & 3-gram & 5.4 & 11.4 & 5.8 & 11.4 & 32.6x \\
~ & 4-gram & 5.4 & 11.3 & 5.7 & 11.3 & 31.5x \\
\hline
\end{tabular}
\label{tab:esa_decode}
\end{table}

Finally, various scoring models are compared. We consider using the AT baseline, neural language model (NLM) and n-gram language model for ranking the sampled alignments. NLM is a transformer-based model trained with the provided text in the LibriSpeech corpus. The n-gram models are also the ones provided in the LibriSpeech corpus. The results of ESA decoding together with that of BPA and ESA decoding are shown in Table \ref{tab:esa_decode}. As analyzed in Sec.\ref{ssec:esa_decoding}, BPA has an impressive speedup but the WER is much worse than the AT baseline, and BSA can obtain a better WER but it is even slower than the AT baseline. By using neural network based scoring models, ESA can retain both the advantages of BPA and BSA. For example, using NLM as a scoring model, ESA can achieve a WER of 8.7\% on the LibriSpeech test-other data and 31.6x speedup. The degradation in speedup compared to BPA originates from the ranking process. Using n-gram models for ranking alignments leads to worse WER compared to NLM in ESA because n-gram models are worse than NLM for language modelling. Another possible reason might be that the probability distribution of n-gram models is  different from that of CASS-NAT decoder outputs, while for NLM it is similar. The reason why n-gram models have similar GPU speedup compared to NLM is that the scores of the sampled alignments cannot be obtained simultaneously. The best performing scoring model is the AT baseline, and thus the AT baseline is used in ESA decoding as default in the following experiments.

Note that because of the sampling process in ESA decoding, WER may be slightly different for different seeds of the random number generator. Fortunately, the randomness is small with a variance of around 0.5\% relatively in our experiments.

%% file: Tex/resultD-F.tex
\subsection{Why does ESA Work? - An Analysis of the CTC Alignments}
\label{ssec:analysis}

In this section, we analyze the CTC alignments obtained from different decoding strategies to understand why ESA can improve the performance of CASS-NAT. 
% Different alignments affect the performance by a having different acoustic range for each token and a different number of predicted tokens for decoder input. Therefore, 
Two metrics are evaluated on the LibriSpeech test sets: mismatch rate (MR) and length prediction error rate (LPER). MR and LPER are measured between the alignments used in decoding and the oracle alignment, in which the blank and repetitive tokens are removed. The MR is the ratio of deletion and insertion errors compared to the oracle alignment, and substitution errors are not included because they do not change either the acoustic boundary for each token or the number of predicted tokens. If the number of tokens in an alignment is different from that in the oracle alignment, this alignment is considered as a case of length prediction error. LPER is the percentage of the utterances with length prediction errors.

\begin{table}[tp]
\caption{\small {Comparisons of different decoding methods for CASS-NAT decoding. \textbf{Oracle}: Viterbi-alignment with ground truth. \textbf{MR}: mismatch rate; \textbf{LPER}: length prediction error rate (using word-piece as the modeling unit). \textbf{S}: the number of alignments sampled in ESA.}}
\footnotesize
\centering
\begin{tabular}{c c cc cc cc}
\hline
\multirow{2}{*}{Decoding} & \multirow{2}{*}{S} & \multicolumn{2}{c}{WER (\%)} & \multicolumn{2}{c}{MR (\%)} & \multicolumn{2}{c}{LPER (\%)} \\
\cmidrule(r){3-4} \cmidrule(r){5-6} \cmidrule(r){7-8} 
Method & ~ & test- & test- & test- & test- & test- & test- \\
~ & ~ & clean & other & clean & other & clean & other \\
\hline \hline
Oracle & 1 & 2.2 & 5.3 & - & - & - & - \\
\hline
BSA & 1 & 4.0 & 9.2 & 2.5 & 6.0  & 28.09 & 48.83 \\ 
BPA & 1 & 4.8 & 10.3 & 2.4 & 5.2 & 34.92 & 51.68 \\
\hline
\multirow{4}{*}{ESA}  & 10  & 4.0 & 9.0 & 3.6 & 6.0 & 26.41 & 44.91 \\
~  & 50  & 3.8 & 8.6 & 3.8 & 6.3 & 25.46 & 43.08 \\
~  & 100 & 3.8 & 8.5 & 3.8 & 6.2 & 25.61 & 42.70 \\
~  & 300 & 3.7 & 8.5 & 3.8 & 6.3 & 25.53 & 42.67 \\
\hline
\end{tabular}
\label{tab:analysis}
\end{table}

\begin{figure}[t]
\centering
\centerline{\includegraphics[width=0.4\textwidth]{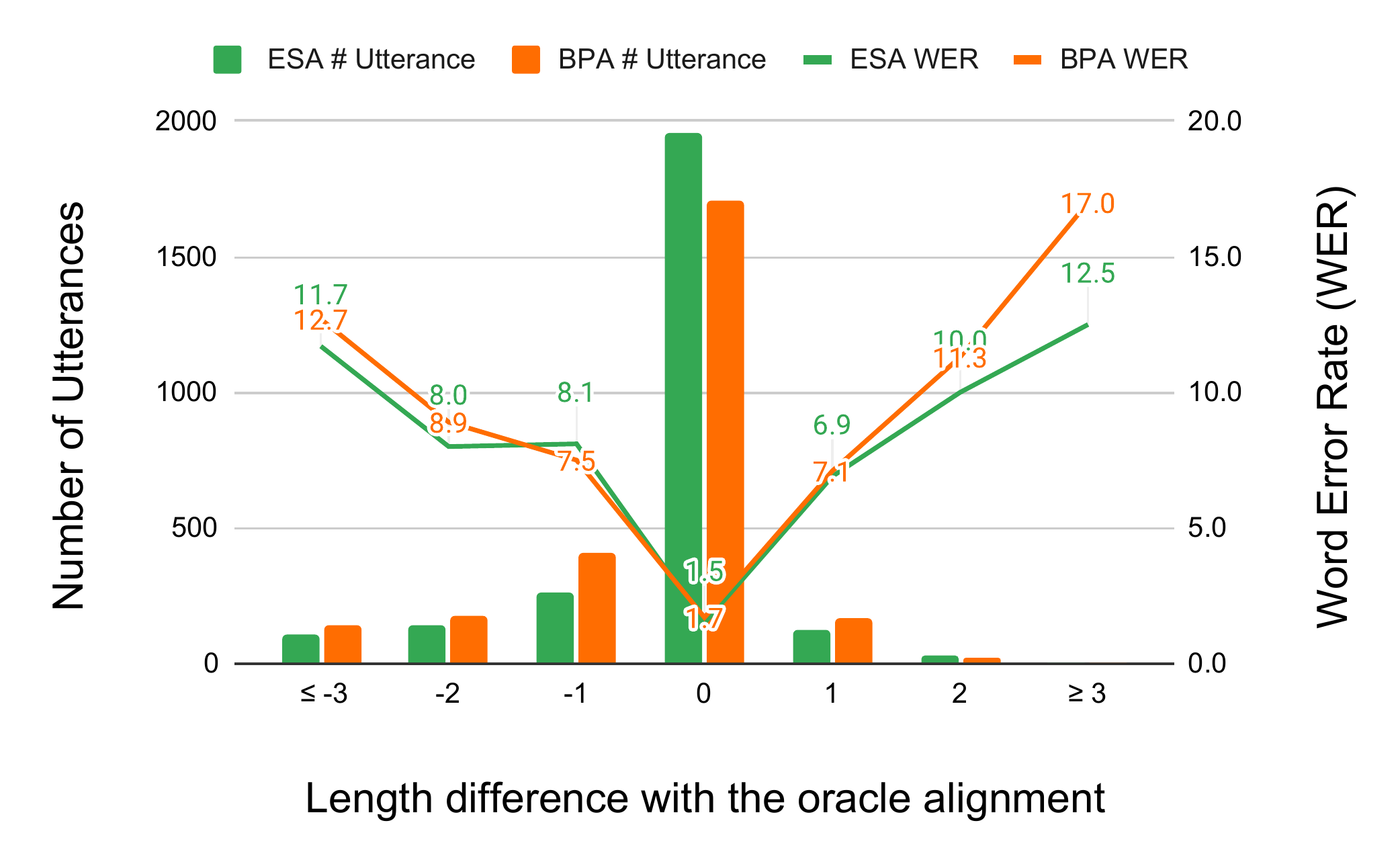}}
\caption{The length prediction error distribution and their corresponding WERs using ESA(s=50) decoding on the test-clean dataset.}
\label{fig:errdist}
\end{figure}

Results of MR and LPER are presented in Table \ref{tab:analysis}. Note that the lower bound of WER (using oracle alignment) is 2.2\% on the test-clean data assuming the transcriptions are available during decoding; this indicates that the framework is promising. When comparing the two metrics for different decoding strategies, we observe that the WER is more correlated with LPER than MR. This suggests that a correct estimation of the output length (the length of decoder input) is very important for NAT, which is also mentioned in \cite{chen2020non}. The reason may be because CASS-NAT decoders have a stronger ability of correcting mistakes in CTC alignment by contextual token-level acoustic modelling when the length prediction is accurate. To further validate this assumption, we plot the length prediction error distribution and their corresponding WERs in Fig.~\ref{fig:errdist}. As seen from the figure, CASS-NAT achieves WERs that are lower than 2\% when the length of the decoder input is estimated correctly, while the WERs are higher than 10\% for those utterances with an absolute difference in length of more than 3 compared to the oracle alignment. 
%Both the results in Table \ref{tab:analysis} and Fig. \ref{fig:errdist} provide a guideline for finding a more effective alignment sampling strategy besides ESA. It also shows the potential of CASS-NAT for obtaining more promising results.

%----------------------------------------------------
%----------------------------------------------------
\subsection{Improving the Training of CASS-NAT}

\begin{table*}[thp]
\caption{WERs of using the proposed training strategies on the LibriSpeech data. SpecAug is used in all configurations. WERR is the relative WER improvement compared to their corresponding baselines on the test-other data. KD represents using knowledge distillation from AT decoder outputs. Other strategies are the ones introduced in Section \ref{ssec:training}. Bold faced numbers indicate the best WER results.}
% NO need -is the ratio of the total inference time to the duration of the test set. LM represents the language model described in Section \ref{sec:expsetup}
%\scriptsize
%\footnotesize
\centering
\begin{tabular}{|c |c|c|c|c|c|c| cccc c|}
\hline
\multirow{2}{*}{Model w/o LM} & \multirow{2}{*}{ConvEnc.} & \multirow{2}{*}{InterCTC} & \multirow{2}{*}{ConvDec.} & \multirow{2}{*}{InterCE} & \multirow{2}{*}{Mask Exp.} &  \multirow{2}{*}{KD} & dev- &  dev- &  test- &  test- & \multirow{2}{*}{\textit{WERR}} \\
~ & ~ & ~ & ~ & ~ & ~ & ~ & clean &  other &  clean &  other & ~ \\
\hline\hline
\rule{0pt}{2ex}
\multirow{3}{*}{AT} & ~ & ~ & ~ & ~ & ~ & ~ & 3.4 & 8.1 & 3.6 & 8.0 & -\\
~ & \Checkmark & ~ & ~ & ~ & ~ & ~ & 2.7 & 6.9 & 2.9 & 6.9 & 13.8\% \\
~ & \Checkmark & \Checkmark & ~ & ~ & ~ & ~ & 2.7 & 6.8 & 2.8 & 6.7 & 16.3\%\\
\hline
\rule{0pt}{2ex}
\multirow{9}{*}{CASS-NAT} & ~ & ~ & ~ & ~ & ~ & ~ & 3.6 & 8.8 & 3.8 & 8.6 & - \\
~ & \Checkmark & ~ & ~ & ~ & ~ & ~ & 3.0 & 7.3 & 3.2 & 7.5 & 12.8\% \\
~ & \Checkmark & \Checkmark & ~ & ~ & ~ & ~ & 2.9 & 7.4 & 3.1 & 7.3 & 15.1\% \\
~ & ~ & ~ & \Checkmark & ~ & ~ & ~ & 3.4 & 8.3 & 3.6 & 8.4 & 2.3\% \\
~ & ~ & ~ & \Checkmark & \Checkmark & ~ & ~ & 3.3 & 8.2 & 3.6 & 8.3 & 3.5\% \\
~ & \Checkmark & \Checkmark & \Checkmark & \Checkmark & ~ & ~ & 2.8 & 7.1 & 2.9 & 7.1 & 17.4\% \\
~ & ~ & ~ & ~ & ~ & \Checkmark & ~ & 3.6 & 8.9 & 3.8 & 8.7 & -1.2\% \\
~ & ~ & ~ & \Checkmark & \Checkmark & \Checkmark & ~ & 3.3 & 8.3 & 3.6 & 8.1 & 5.8\% \\
~ & \Checkmark & \Checkmark & \Checkmark & \Checkmark & \Checkmark & ~ & \textbf{2.7} & \textbf{7.1} & \textbf{2.9} & \textbf{7.0} & 18.6\% \\
~ & ~ & ~ & ~ & ~ & ~ & \Checkmark & 3.6 & 8.8 & 3.8 & 8.5 & 1.2\% \\

\hline
\end{tabular}
\label{tab:improvements_libri}
\end{table*}

Despite the use of ESA decoding method, the WER performance of CASS-NAT (8.6\%) is still worse than that of the AT baseline (8.0\%). In this section, we attempt to improve the WER performance of CASS-NAT by using various training strategies that include convolution-augmented encoder (ConvEnc.), intermediate CTC loss (InterCTC), convolution augmented decoder (ConvDec.), intermediate CE loss (InterCE) and trigger mask expansion (Mask Exp.). The results of various combinations of the training strategies are shown in Table \ref{tab:improvements_libri}. 

First, we apply ConvEnc. and InterCTC to the autoregressive transformer to create a new AT baseline for fair comparisons. Note that, we have applied ConvDec. and InterCE to the AT baseline as well, but no improvements are observed, and thus their results are not included. When ConvEnc. and InterCTC are applied to CASS-NAT, similar improvements compared to their use for AT are observed. The improvements stem from their ability of capturing local details in the acoustic representations, and thus a stronger encoder is learned and it offers an accurate CTC alignment for implicit token-level acoustic embedding modelling in the decoder. 

Second, when using ConvDec. and InterCE, the WER of CASS-NAT drops further to 7.1\% on the test-other data, which is a 17.4\% relative WER improvement over the CASS-NAT baseline. The use of ConvDec. and InterCE, however, is not as effective as ConvEnc and InterCTC, indicating that a stronger modelling of frame-level acoustic representations is more important than that of token-level acoustic representations (TAEs) in CASS-NAT. This is reasonable because TAEs are extracted from frame-level acoustic representations from the encoder. Interestingly, we observe a -1.2\% relative WER reduction when the trigger mask expansion method is applied by itself. However, the method leads to WER improvements when it is used together with ConvDec. related strategies. The reason could be that expanding the acoustic boundary for each token might cause confusion for the decoder when the contextual modelling is not strong enough. 

In addition, as shown in the last line of Table \ref{tab:improvements_libri}, applying knowledge distillation, which takes the AT output as a target for NAT training, does not result in WER improvements. Although knowledge distillation has been proven useful in non-autoregressive neural machine translation in \cite{zhou2020understanding}, it does not seem to work for speech recognition possibly because there are no different outputs corresponding to the same input, while multiple translations of the source language exist in the target language for machine translation. 

Finally, using all the training strategies, we achieve an 18\% relative WER improvement compared to CASS-NAT baseline; this is only a 3\% WER degradation compared to the best version of the AT baseline. In summary, we obtain a non-autoregressive speech transformer that performs close to its autoregressive counterpart with a significant GPU speedup.  

\begin{figure}[t!]
    \centering
    \subfloat[The $5^{th}\sim8^{th}$ head of self-attention in the last SAD]{
    {\includegraphics[width=0.115\textwidth]{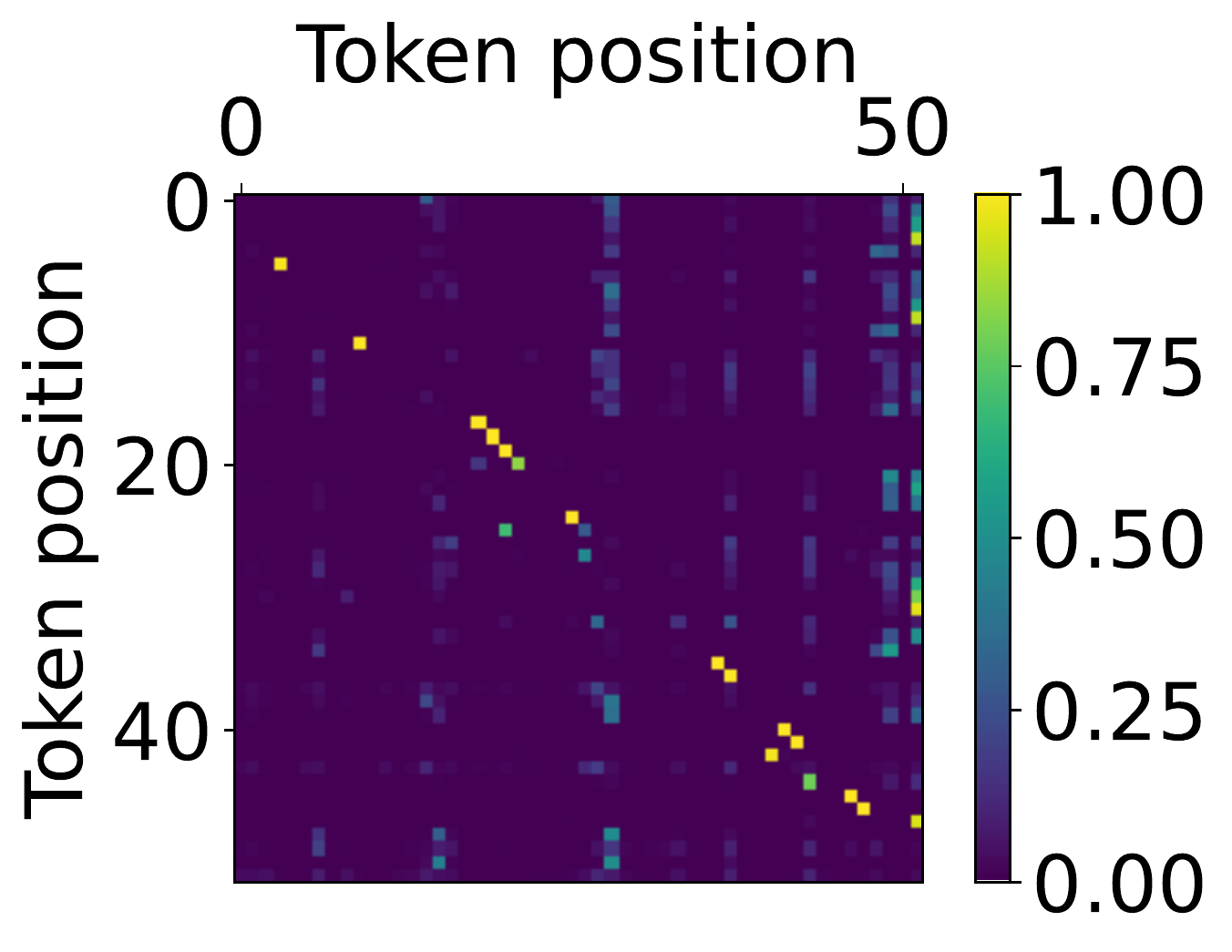}} %
    {\includegraphics[width=0.115\textwidth]{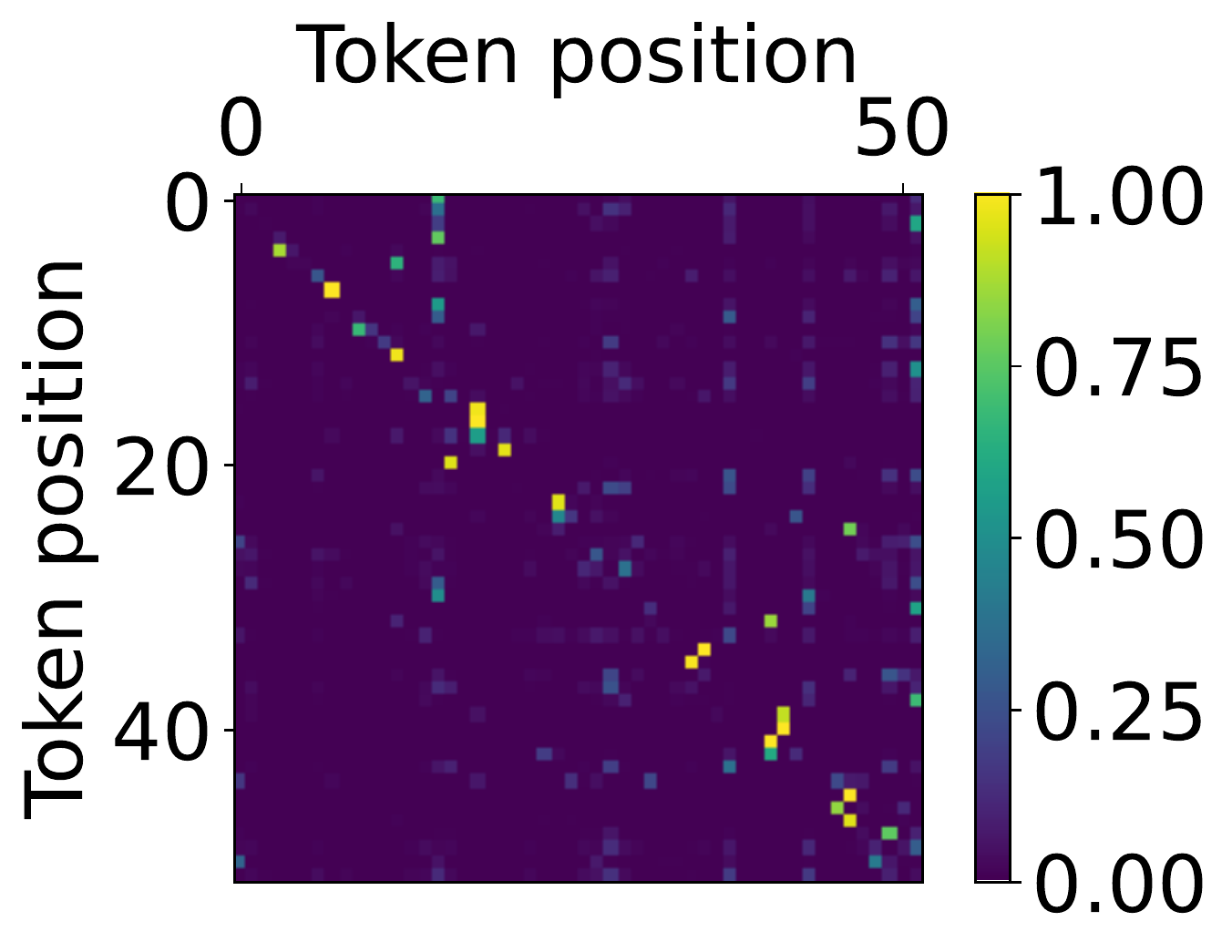} }%
    \includegraphics[width=0.115\textwidth]{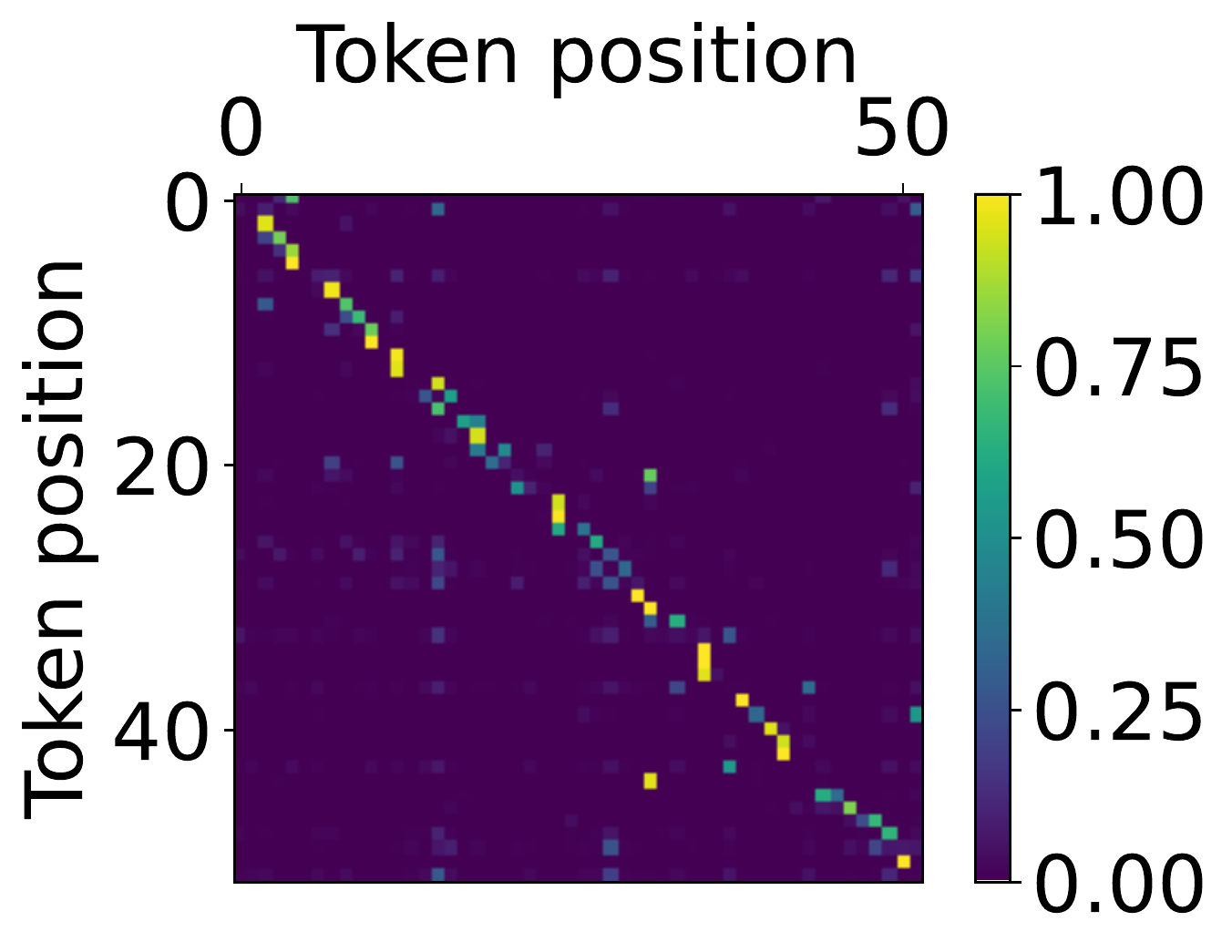} %
    \includegraphics[width=0.115\textwidth]{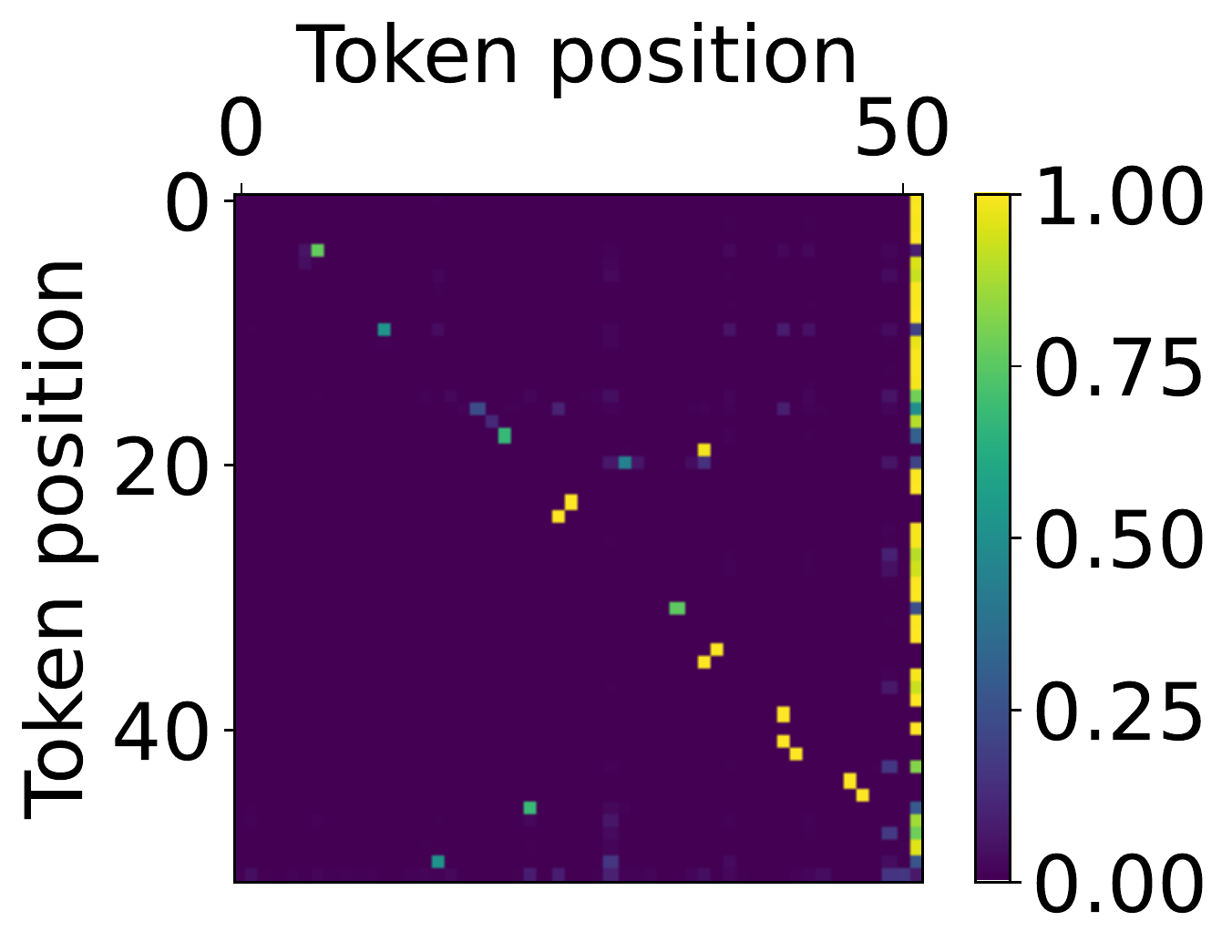} } %

    \subfloat[The $5^{th}\sim8^{th}$ head of self-attention in the last MAD]{
    {\includegraphics[width=0.115\textwidth]{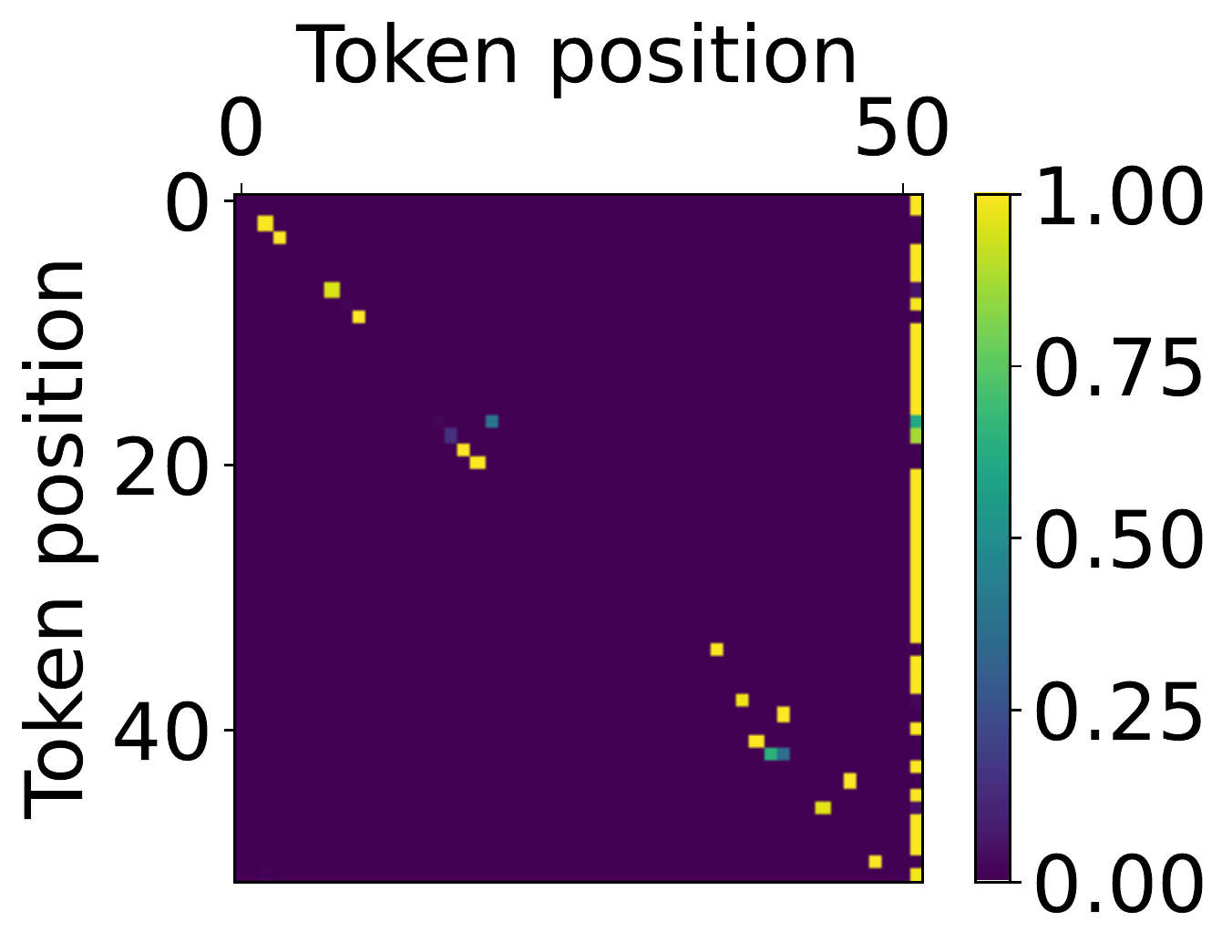}} %
    {\includegraphics[width=0.115\textwidth]{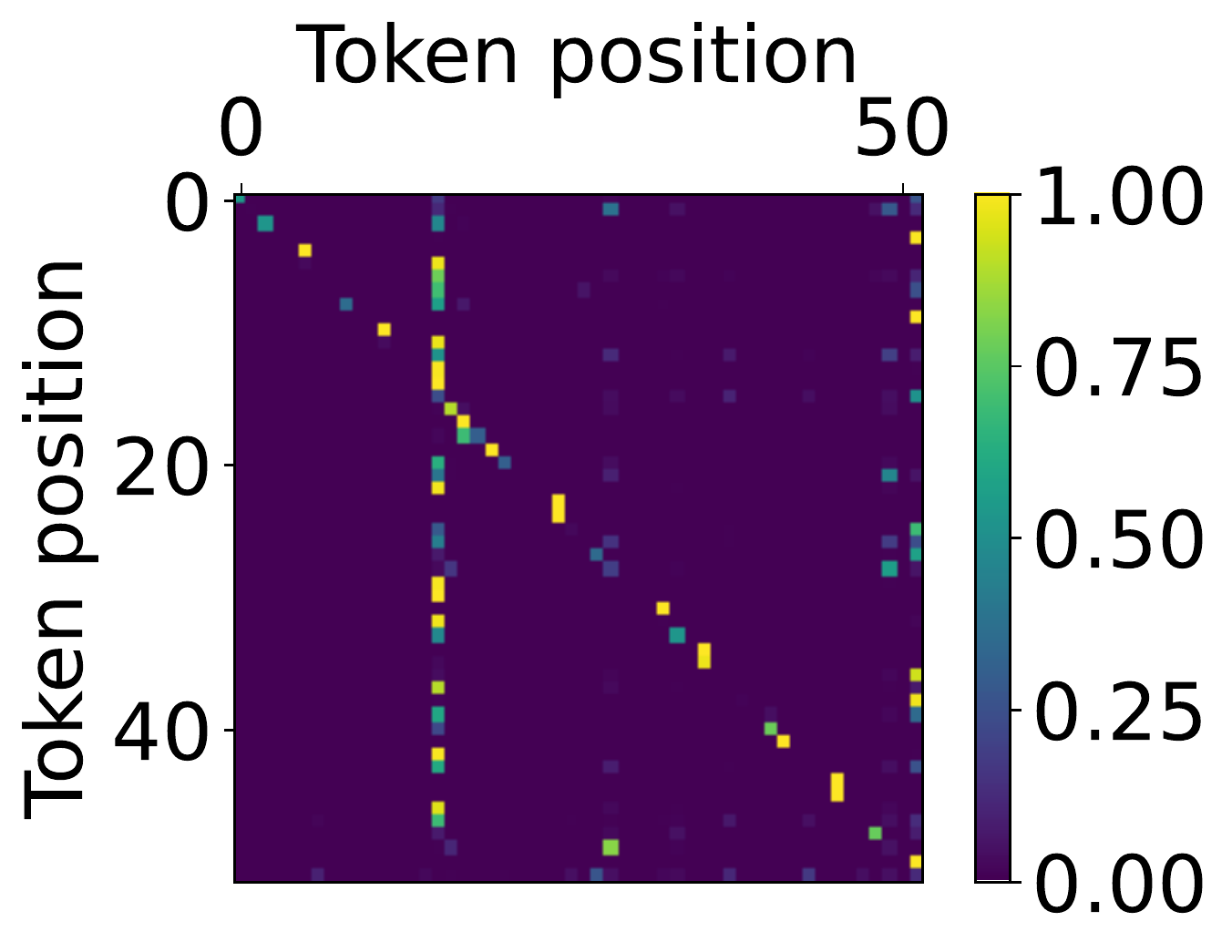}} %
    {\includegraphics[width=0.115\textwidth]{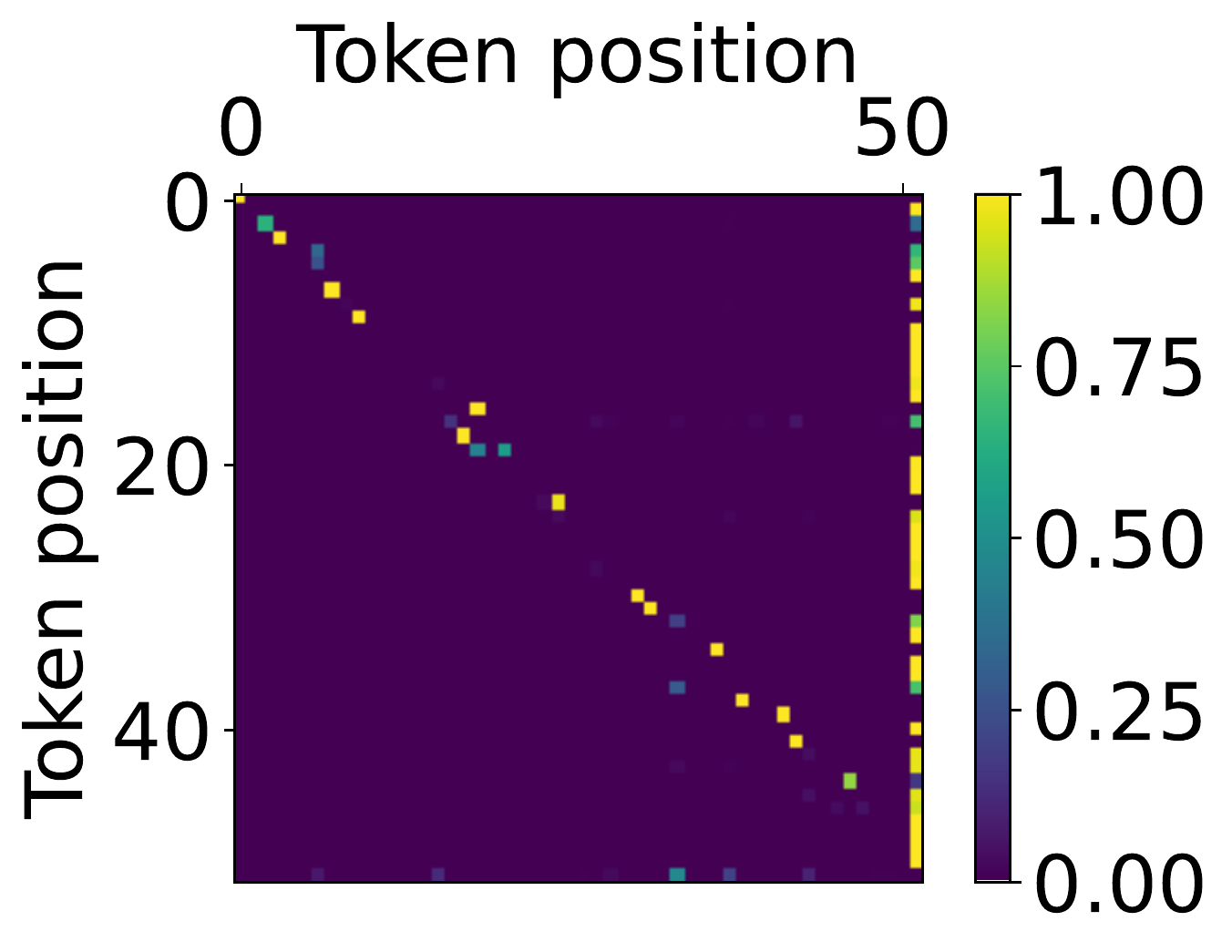}}  %
    {\includegraphics[width=0.115\textwidth]{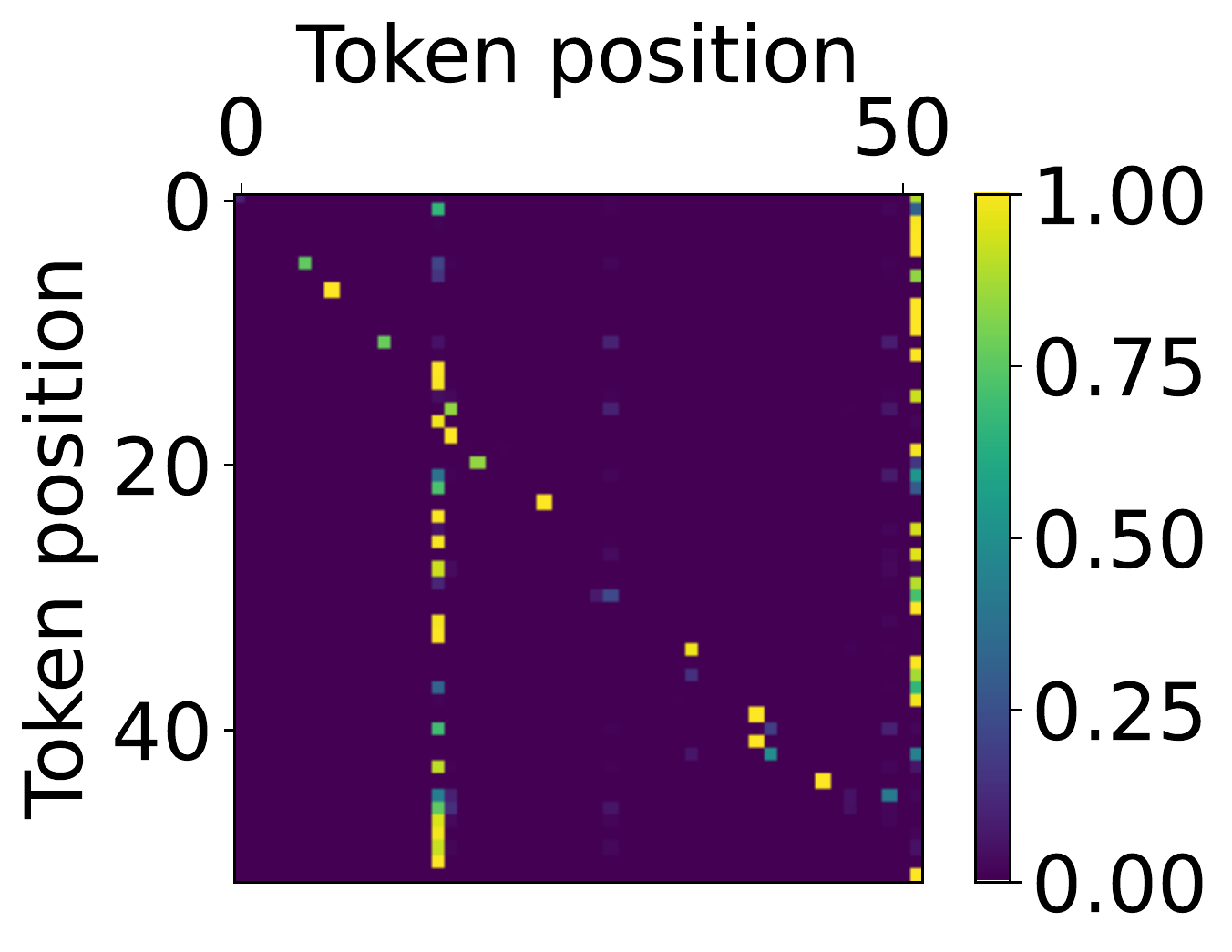}}} %
    \caption{Attention weight distributions of the $5^{th}-8^{th}$ head in the last self-attention decoder (SAD) and mixed-attention decoder (MAD). The first utterance in the LiriSpeech train-clean-100 subset is used. The y-axis represents output tokens} %
    \label{fig:attweight}%
\end{figure}
%----------------------------------------------------
%----------------------------------------------------

%\subsection{Effect of Encoder Initialization}
\subsection{Why does CASS-NAT Work? - An Analysis of the Decoder}
\label{ssec:why_cassnat}

\begin{figure*}[t!]
    \centering
    \subfloat[TAEE output]{{\includegraphics[width=0.3\textwidth]{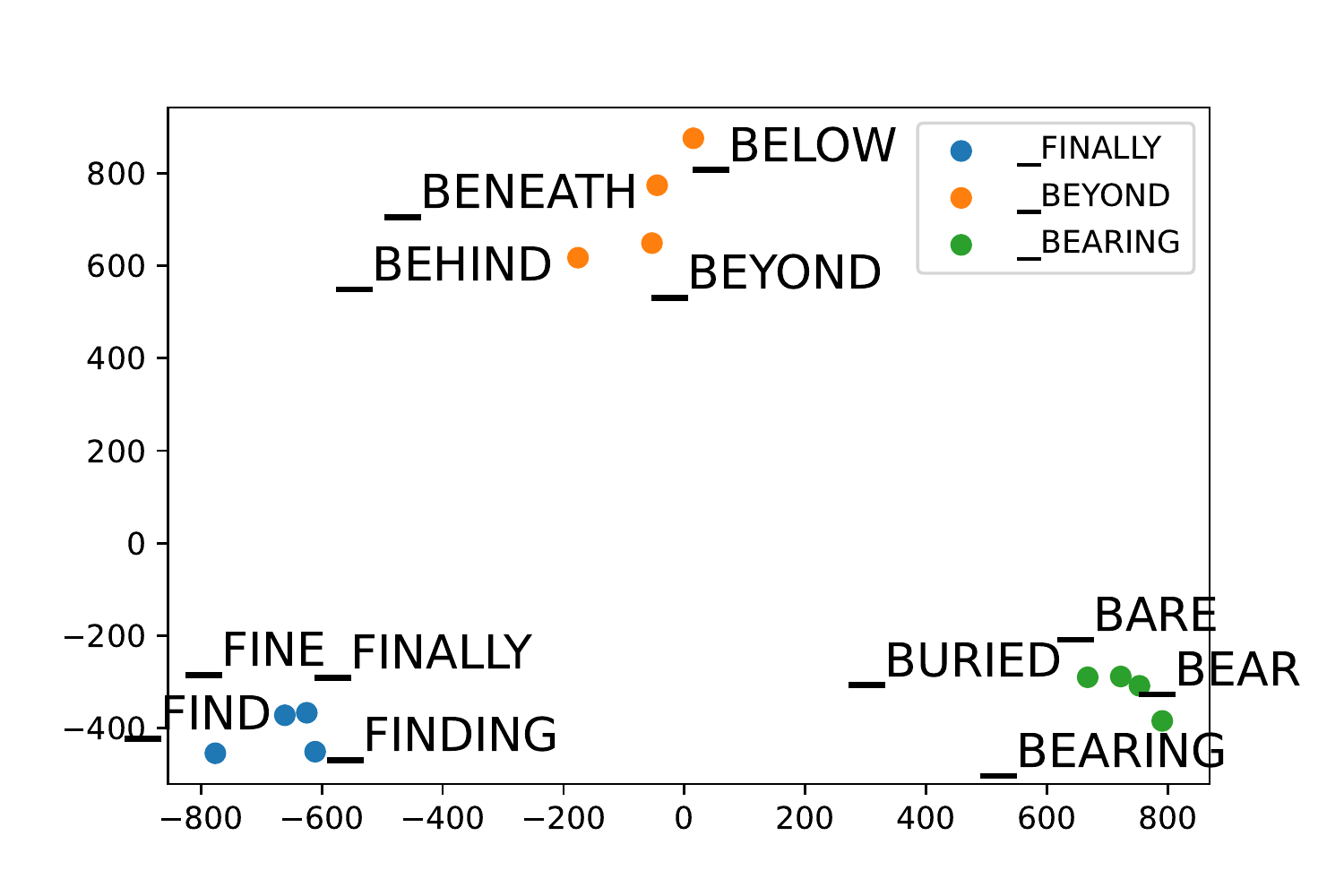} }}%
    %\qquad
    \subfloat[SAD output]{{\includegraphics[width=0.3\textwidth]{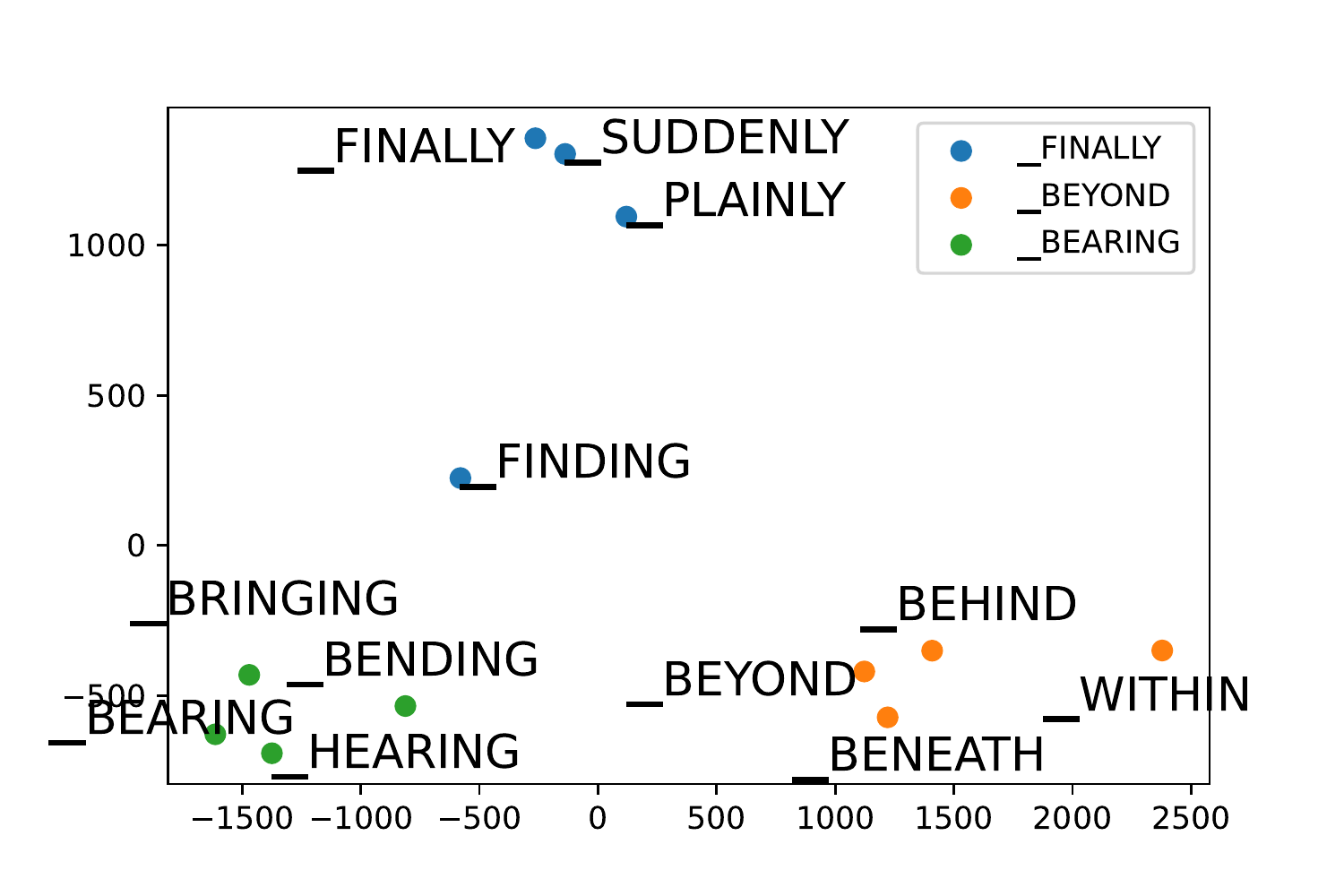}}}%
     \subfloat[MAD output]{{\includegraphics[width=0.3\textwidth]{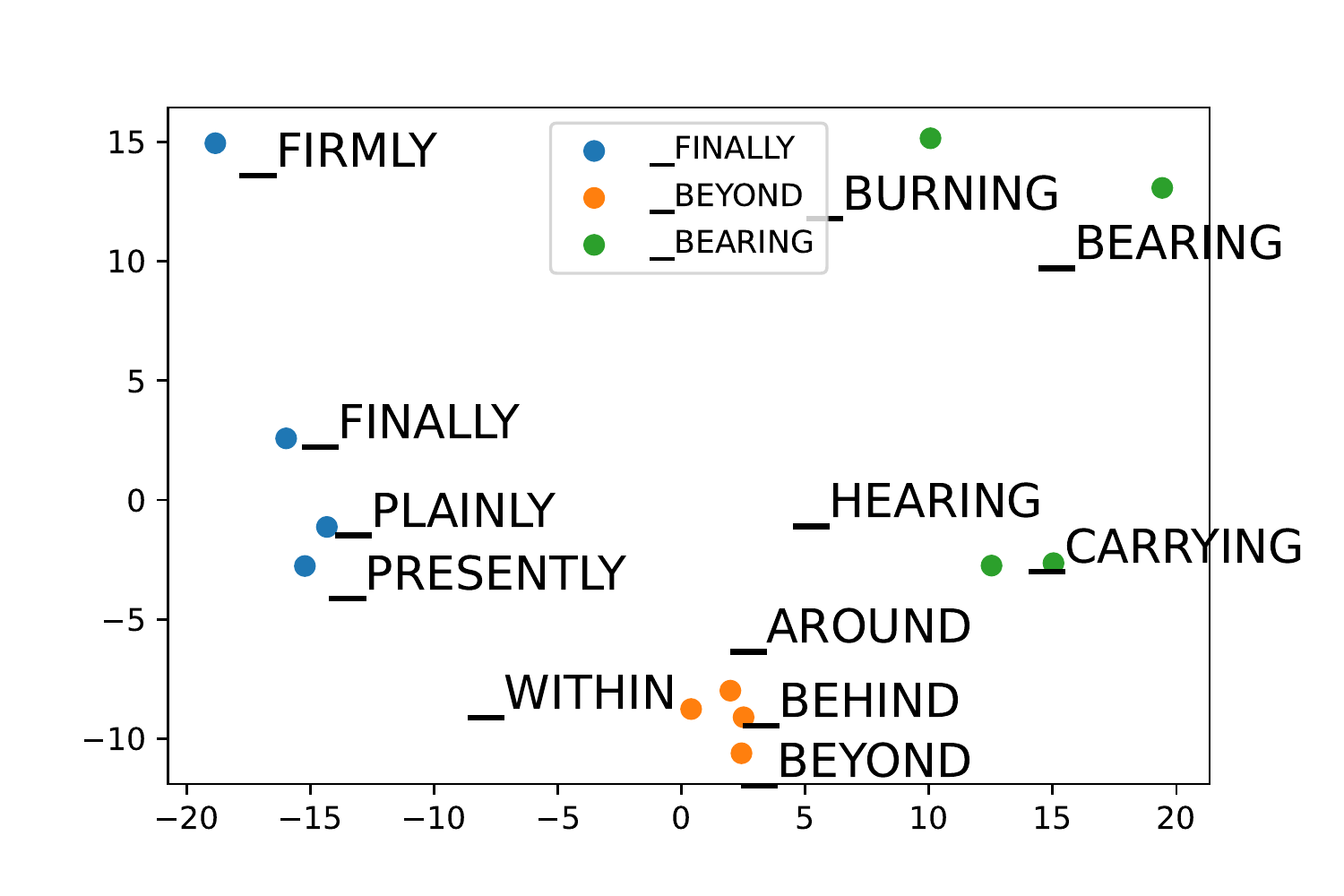}}}%
    \caption{Visualization of token-level acoustic embeddings of three word pieces from outputs of TAEE, SAD and MAD, respectively. The embeddings are plotted using principle component analysis (PCA).}%
    \label{fig:acembed}
\end{figure*}

In this section, we explain why CASS-NAT can achieve a good performance that is close to its autoregressive counterpart by analysing the following elements in CASS-NAT decoder: (1) attention weight distribution; and (2) acoustic-level acoustic representations.

First, to analyse the behaviour of the attention mechanism in CASS-NAT decoder, we choose the first utterance in the LibriSpeech train-clean-100 subset and plot the attention weight distribution in the last SAD and MAD blocks. The weights of the last 4 heads (8 heads in total) are shown in Fig.\ref{fig:attweight}. For self-attention weight distributions in SAD, we notice from the figure that most of the heads learn a monotonic alignment between token-level acoustic representations, indicating that each token relies on adjacent tokens. This is similar to the idea of word embedding using a continuous bag of words (CBOW) and skip-gram \cite{mikolov2013efficient}. The monotonic alignment also shows the usefulness of the relative positional encoding because distant tokens with close semantic similarity have low attention weights. For the attention weights in MAD, there exists a different behaviour in several subspaces of the attention computation, where outputs may rely on the same input (vertical line in Fig.\ref{fig:attweight}b).

\begin{table*}[thp]
\caption{WER Results of CASS-NAT using different encoders as a starting point, including random initialization (Rand. Enc.), initialization from an encoder trained with CTC (CTC enc.), and initialization from a AT encoder (AT Enc.). AT baselines with greedy and beam search decoding are included together with SOTA results using non-autoregressive methods in the literature. ``-" indicates that WER results have not been reported in the literature. RTF is evaluated with batch size of one on GPU. Bold-faced numbers are the best results for CASS-NAT and SOTA without self-supervised pretraining.}

\centering
\begin{tabular}{c c cc cccc  cc cc cc}
\hline
\rule{0pt}{2ex}
\multirow{3}{*}{Model} & \multirow{3}{*}{Conditions} & 
\multicolumn{2}{c}{Inference Speed} & \multicolumn{4}{c}{LibriSpeech} & \multicolumn{2}{c}{Aishell1} & \multicolumn{2}{c}{TED2} & \multicolumn{2}{c}{MyST} \\
\cmidrule(r){3-4} \cmidrule(r){5-8} \cmidrule(r){9-10} \cmidrule(r){11-12} \cmidrule(r){13-14}
~ & ~ & \multirow{2}{*}{RTF} & \multirow{2}{*}{Speed} & \multicolumn{2}{c}{dev} & \multicolumn{2}{c}{test} & \multirow{2}{*}{dev} & \multirow{2}{*}{test} & \multirow{2}{*}{dev} & \multirow{2}{*}{test} & \multirow{2}{*}{dev} & \multirow{2}{*}{test} \\
~ & ~ & ~ & ~ & clean & other & clean & other & ~ & ~ & ~ & ~ & ~ & ~ \\
\hline\hline
\rule{0pt}{2ex}
\multirow{2}{*}{AT} & greedy search &  0.0387 & 1.00x & 2.7 & 7.0 & 2.9 & 7.0 & 5.1 & 5.9 & 9.8 & 7.8 & 18.1 & 17.0 \\
~ & beam search & 0.3348 & 0.12x & 2.7 & 6.8 & 2.8 & 6.7 & 4.7 & 5.1 & 8.2 & 7.6 & 16.5 & 16.2 \\
\hdashline
\rule{0pt}{2ex}
\multirow{2}{*}{\parbox{1.5cm}{Previous NAT results}} & w/o SSL & ~ & ~ & 2.8 & 7.3 & 3.1 & 7.2\textsuperscript{\cite{fan2021improved}} & 4.5 & 4.9 \textsuperscript{\cite{yu2021boundary}} & 8.7 & 8.0\textsuperscript{\cite{higuchi2021comparative}} & 28.0 & 27.8\textsuperscript{\cite{9864219}} \\

~ & w/ SSL & ~ & ~ & 1.7 & 3.6 & 1.8 & 3.6\textsuperscript{\cite{ng2021pushing}} & 4.1 & 4.5 \textsuperscript{\cite{deng2022improving}} & - & - & 16.8 & 16.5\textsuperscript{\cite{interspeech/FanA22}}\\
\hdashline
\rule{0pt}{2ex}
\multirow{3}{*}{\parbox{1.5cm}{CASS-NAT (this work)}} & Rand. Enc. & \multirow{3}{*}{0.0134} & \multirow{3}{*}{2.89x} & 3.3 & 8.4 & 3.6 & 8.3 & \textbf{4.7} & \textbf{5.0} & 8.4 & \textbf{7.5} & \textbf{16.8} & \textbf{16.4} \\
~ & CTC Enc. & ~ & ~ & 2.8 & 7.2 & 3.1 & 7.3 & 5.2 & 5.6 & \textbf{8.0} & 7.7 & 17.1 & 16.7  \\
~ & AT Enc. & ~ & ~ & \textbf{2.7} & \textbf{7.1} & \textbf{2.9} & \textbf{7.0} & 5.1 & 5.4 & 8.2 & 7.8 & 17.1 & 16.7 \\

~ & HuBERT Enc. & ~ & ~ & - & - & - & - & 4.2 & 4.5 & - & - & 16.0 & 15.6  \\

\hline
\end{tabular}
\label{tab:all_results}
\end{table*}

As discussed in Section \ref{sssec:ss_nat}, we assume that language semantics can be captured by CASS-NAT decoder using token-level acoustic embeddings. To validate this assumption, we calculate three different token-level acoustic embeddings from TAEE, SAD and MAD for each token (word piece in our case). Specifically, token-level acoustic representations are first extracted from the first 5000 utterances in the train-clean-100 Librspeech subset, where each representation has its label in the form of a word piece. The embedding of each word piece is the average of the corresponding representations. After that, we randomly choose three word pieces and find the four closest ones (measured with cosine similarity distance) for each word piece. Using the same idea of visualizing word embeddings, the 12 selected embeddings are reduced to a 2-dimensional space using principal component analyses (PCA) and are then plotted. Examples of embeddings at different levels of the CASS-NAT decoder are shown in Fig.\ref{fig:acembed}. The figure suggests that the token-level acoustic embeddings learn not only acoustic similarities (\textbf{\_BEAR} vs. \textbf{\_BARE}), but also token-level semantic similarities (\textbf{\_BENEATH} vs. \textbf{\_BELOW}). This occurs, even though  embeddings at different levels of the decoder provide different information. For example, higher layer (MAD) embeddings focus more on grammatical similarities (e.g. \textbf{\_FIRMLY} and \textbf{\_FINALLY}), which is similar to word embedding. This finding suggests the possibility of learning a joint speech and text embeddings in a common space. Even though there is no explicit language modelling, the CASS-NAT decoder is able to learn meaningful embeddings, which may explain why it has a similar performance to its AT counterpart.

\subsection{Results on other datasets}
\label{ssec:all_results}
Finally, we apply the proposed CASS-NAT to the other three datasets: Aishell (Mandarin speech), TED2 (English spontaneous speech), and MyST (child speech), to show the generalization ability of CASS-NAT. In addition, we explore the effect of different initialization schemes for the CASS-NAT encoder and measure the speedup given by CASS-NAT compared to AT baselines. All results are summarized in Table \ref{tab:all_results}. As can be observed from the table, a randomly initialized encoder for CASS-NAT achieves the best performance on the Aishell1, TED2 and MyST datasets, while the LibriSpeech dataset performs better with a pretrained encoder. The reason could be that we force the training iterations of CASS-NAT to be the same as the AT baseline for time considerations, as well as for a fair comparison, but LibriSpeech, as a large database compared to the others, needs more training iterations to obtain accurate alignments for the decoder. Hence, we suggest using a pretrained encoder for training CASS-NAT on large datasets.

With an appropriate encoder initialization, we now compare the results of CASS-NAT to the AT baselines and SOTA results using NATs. First, when compared to the AT baseline with greedy search decoding, CASS-NAT achieves better (on Aishell1, TED2, and MyST) or similar (on LibriSpeech) WERs performance, and has a 2.89x RTF speedup. When compared to the AT baseline with beam search decoding, the WERs performance on Aisehll and TED2 are still better, while the WERs on LibriSpeech and MyST are close to the baseline but with a 24x RTF speedup. Second, when compared to other NAT methods in the literature, we observe from the table that the proposed CASS-NAT achieves new state-of-the-art results on LibriSpeech, TED2, and MyST when no pretrained acoustic (e.g. Wav2vec2.0 \cite{baevski2020wav2vec}) or language models (e.g. BERT\cite{devlin2018bert}) are used. To compare with the results that used extra training data (e.g. SSL with un-annotated data), we use the self-supervised pretrained models as the encoder for CASS-NAT to see whether it can outperform the SOTA results in Table \ref{tab:all_results}. Due to time and computational constraints, we obtain only the results on Aishell1 and MyST datasets. The results show that CASS-NAT can achieve comparable results to the best NATs in the literature \cite{deng2022improving,interspeech/FanA22} with SSL pretraining.

%% file: Tex/conclusion.tex
\section{Conclusions}
\label{sec:conclusion}
This paper presents a comprehensive study of the CASS-NAT framework introduced in \cite{fan2021cass}, which utilizes alignments obtained from the CTC output space to extract token-level acoustic embeddings (TAEs), and regards the TAEs as substitutions of the word embeddings in autoregressive transformers (AT) to achieve parallel computation. During training, Viterbi-alignment is used to estimate the posterior probability, and various training strategies are used to improve the WER performance of the CASS-NAT. During inference, we investigate in depth an error-based sampling alignment (ESA) method that we introduced recently in \cite{fan2021cass} to reduce the alignment mismatch between training and inference. Detailed experiments show that: (1) a mixed-attention decoder (MAD) is important for reducing the WER; (2) the reason why ESA decoding works well is because it has a lower length prediction error rate; (3) convolution augmented encoder and decoder, intermediate loss and mask expansion can improve the WER of CASS-NAT, while knowledge distillation can not; and (4) TAEs have similar functionality to word embeddings, such as representing grammatical structures, indicating the possibility of learning semantics without a language model. As a result, CASS-NAT achieves new state-of-the-art results for NAT models on several ASR tasks with and without SSL pretraining. The datasets include English and Mandarin speech, read and spontaneous speech, and child and adult speech, showing the generaliztion ability of the proposed method. The performances of CASS-NAT are comparable, in relative terms, to AT with beam search decoding, but maintain a $\sim$24x speed up. Future work includes investigating the incorporation of language models using external data.